\documentclass[conference]{IEEEtran}
\IEEEoverridecommandlockouts
\usepackage{cite}
\usepackage{amsmath,amssymb,amsfonts}
\usepackage{algorithmic}
\usepackage{graphicx}
\usepackage{grffile}
\usepackage{textcomp}
\usepackage{xcolor}

\usepackage{multirow}
\usepackage{multicol}
\usepackage{booktabs}       
\usepackage{xspace}
\usepackage{float}
\usepackage{xcolor}
\usepackage{comment}
\usepackage{amsmath}
\usepackage{bm}
\usepackage{enumitem}
\usepackage{nicematrix}
\usepackage[export]{adjustbox}
\usepackage{subcaption}
\usepackage[capitalize]{cleveref}
\usepackage[misc]{ifsym}
\usepackage{url}

\Crefname{equation}{Eq.}{Eqs.}
\Crefname{Figure}{Fig.}{Figs.}

\newcommand{\revise}[1]{\textcolor{black}{#1}}

\newcommand{\ourmodel}{{Conformer}\xspace}

\newcommand{\defeq}{\overset{\underset{\mathrm{def}}{}}{=}}

\newcommand{\eat}[1]{}
\newcommand{\ie}{\emph{i.e.,}\xspace}
\newcommand{\eg}{\emph{e.g.,}\xspace}

\def\BibTeX{{\rm B\kern-.05em{\sc i\kern-.025em b}\kern-.08em
    T\kern-.1667em\lower.7ex\hbox{E}\kern-.125emX}}

\usepackage{fancyhdr}
\begin{document}

\title{
Towards Long-Term Time-Series Forecasting: Feature, Pattern, and Distribution
}

\author{
\IEEEauthorblockN{
Yan Li%
\textsuperscript{$\S \dagger  \ast$}%
\thanks{
{\textsuperscript{$\ast$}}%
This work was done when the first author was an intern at 
Baidu Research under the supervision of 
the second author.
},
Xinjiang~Lu%
\textsuperscript{$\dagger \, {\textrm{\Letter}}$},
Haoyi~Xiong%
\textsuperscript{$\dagger$},
Jian~Tang%
\textsuperscript{$\P \, \natural$}, 
Jiantao~Su%
\textsuperscript{$\natural$}, 
Bo~Jin%
\textsuperscript{$\sharp$}, 
Dejing~Dou%
\textsuperscript{$\dagger$}
}
\IEEEauthorblockA{
\textit{
\textsuperscript{$\dagger$}%
Baidu Research, 
\textsuperscript{$\S$}%
Zhejiang University, 
\textsuperscript{$ \P $}%
Tsinghua University, }\\
\textit{
\textsuperscript{$ \natural $}%
China Longyuan Power Group Corp. Ltd.,
\textsuperscript{$ \sharp $}%
Dalian University of Technology
}\\
ly21121@zju.edu.cn,
$\{$luxinjiang,xionghaoyi$\} \text{@baidu.com}$,
$\{$12101779,12091329$\} \text{@chnenergy.com.cn}$, \\
jinbo@dlut.edu.cn, dejingdou@gmail.com
}}

\maketitle
\thispagestyle{fancy}

\begin{abstract}
Long-term time-series forecasting (LTTF) has become a pressing demand in many applications, such as wind power supply planning. 
Transformer models have been adopted to deliver high prediction capacity because of the high computational self-attention mechanism.
Though one could lower the complexity of Transformers by inducing the sparsity in point-wise self-attentions for LTTF,
the limited information utilization prohibits the model from exploring the complex dependencies comprehensively. 
To this end, we propose an efficient Transformer-based model, named \ourmodel, which differentiates itself from existing methods for LTTF in three aspects: 
(i) an encoder-decoder architecture incorporating a linear complexity without sacrificing information utilization  is proposed on top of sliding-window attention and Stationary and Instant Recurrent Network (SIRN); 
(ii) a module derived from the {\em{normalizing flow}} is devised to further improve the information utilization by inferring the outputs with the latent variables in SIRN directly; 
(iii) the inter-series correlation and temporal dynamics in time-series data are modeled explicitly to fuel the downstream self-attention mechanism. 
Extensive experiments on seven real-world datasets demonstrate that \ourmodel outperforms the state-of-the-art methods on LTTF and generates reliable prediction results with uncertainty quantification.  
\end{abstract}

\begin{IEEEkeywords}
Long-term time-series forecasting, Transformer, Normalizing Flow
\end{IEEEkeywords}

\section{Introduction}

Time-series data evolve over time, which can result in perplexing time evolution patterns over the short- and long-term.
The time evolution nature of time-series data is of great interest to many downstream tasks including time-series classification, outlier detection, and time-series forecasting.
Among these tasks, time-series forecasting (TF) has attracted many researchers and practitioners in a wide range of application domains, 
such as transportation and urban planning~\cite{lstnet}, 
energy and smart grid management~\cite{wang2021review}, 
as well as weather~\cite{han2021joint} and disease propagation analysis~\cite{matsubara2014funnel}.

In many real-world application scenarios, given a substantial amount of time-series data recorded, there is a necessity to make a decision in advance, such that, with long-term prediction, the benefits can be maximized while the potential risks can be avoided.
Therefore, in this work, we study the problem of forecasting time series that looks far into the future, namely long-term time-series forecasting (LTTF).

While tons of TF methods~\cite{arima,var,evar,svr} have been proposed with statistical learners, the use of domain knowledge however seems indispensable to model the temporal dependencies for TF but also limits the potential in applications. 
Recently,  deep models~\cite{deepar,grnn,sscnn,reformer,autoformer} have been proposed for TF, which can be categorized into two types: the RNN-based and the Transformer-based models.
RNN-based methods capture and utilize long- and short-term temporal dependencies to make the prediction, but fail to deliver good performance in long-term time-series forecasting tasks. 
Transformer-based models have achieved promising performance in extracting temporal patterns for LTTF because of the usage of self-attention mechanisms.
However, such ``full'' attention mechanisms bring quadratic computation complexity for TF tasks, which thus becomes the main bottleneck for Transformer-based models to solve the long-term time-series forecasting task.

Several works have been devoted to improving the computation efficiency of self-attention mechanisms and lowering the complexity of handling a length-$L$ sequence to ($\mathcal{ O } (L \log {L})$ or $ \mathcal{O} (L  \sqrt{L}) $), such as Logtrans~\cite{logsparse}, Reformer~\cite{reformer}, Informer~\cite{Informer} and Autoformer~\cite{autoformer}. 
In the NLP field, some pioneering works have been proposed to reduce the complexity of self-attention to linear ($ \mathcal{O} (L) $), including Longformer~\cite{longformer} and BigBird~\cite{bigbird}. 
However, {{these deep models with a linear complexity might}} limit the information utilization and  strain the performance of LTTF. 
Lowering the computational complexity to $\mathcal{O} (L)$ without sacrificing information utilization is a big challenge for LTTF.

In addition to the complexity, as the input length climbs up, the intricate time-series could exhibit obscure and confusing temporal patterns, which may lead to unstable prediction for self-attention-based models. 
Moreover, multivariate long-term time-series often embody multiple temporal patterns at different temporal resolutions, \eg seconds, minutes, hours, or days. 
On the other hand, the intricate and prevailing multi-dimensional characteristics of the time-series data exhibit multi-faceted complex correlations among different series. 
Therefore, how to make the prediction for LTTF more stable and disaggregate multiscale dynamics and multivariate dependencies in time-series data are two more challenges.

To this end, our work devotes to the above three challenges and proposes a novel model based on Transformer for LTTF, namely  \ourmodel. 
In particular, \ourmodel first explicitly explores the inter-series correlations and temporal dependencies with Fast Fourier Transform (FFT) plus multiscale dynamics extraction. 
Then, to address the LTTF problem in a sequence-to-sequence manner with linear computational complexity, an encoder-decoder architecture is employed on top of the sliding-window self-attention mechanism and the proposed stationary and instant recurrent network (namely, SIRN). 
More specifically, the sliding-window attention allows each point to attend to its local neighbors for reference, such that the self-attention dedicated to a length-$L$ time-series requires the $\mathcal{O} (L)$ complexity. 
Besides, to explore global signals in time-series data without violating {{the linear complexity}}, we renovate the cycle structure of the recurrent neural network (RNN) and distill stationary and instant patterns in long-term time-series with {{the}} series decomposition model in a recurrent way.

Moreover, to relieve the fluctuation effect caused by the {aleatoric uncertainty~\cite{gawlikowski2021survey}} of {{time series data}} and improve the prediction reliability for LTTF, we further put efforts to model the underlying distribution of {{time-series data}}. 
To be specific,  we devise a {\em{normalizing flow}} block to absorb {latent states} yielded in the SIRN model and generate the distribution of {{future series}} directly. 
More specifically, we leverage the outcome latent state of the encoder, as well as the latent state of the decoder, as input to initiate the normalizing flow. 
Afterward, the latent state of the decoder can be cascaded to infer the distribution of the target series. 
Along this line, the information utilization for LTTF can be further enhanced and the time-series forecasting can be implemented in a generative fashion, which is more noise-resistant.

Extensive experiments on seven real-world datasets validate that \ourmodel outperforms the state-of-the-art {{(SOTA)}} baselines with satisfactory margins.
To sum up, our contributions can be highlighted as follows:

\begin{enumerate}[leftmargin=0.5cm,label={$\bullet$}]

\item
We reduce the complexity of self-attention to $\mathcal{O}(L)$ without sacrificing prediction capacity with the help of windowed attention and the renovated recurrent network.

\item
We design a normalizing flow block to infer target series from hidden states directly, which can further improve the prediction and equip the output with uncertainty awareness. 

\item 
{{Extensive experiments on five benchmark datasets and two collected}} datasets validate the superior long-term time-series forecasting performance of \ourmodel. 

\end{enumerate}

\begin{figure*}[t]
\centering
  \includegraphics[width=0.9\textwidth]{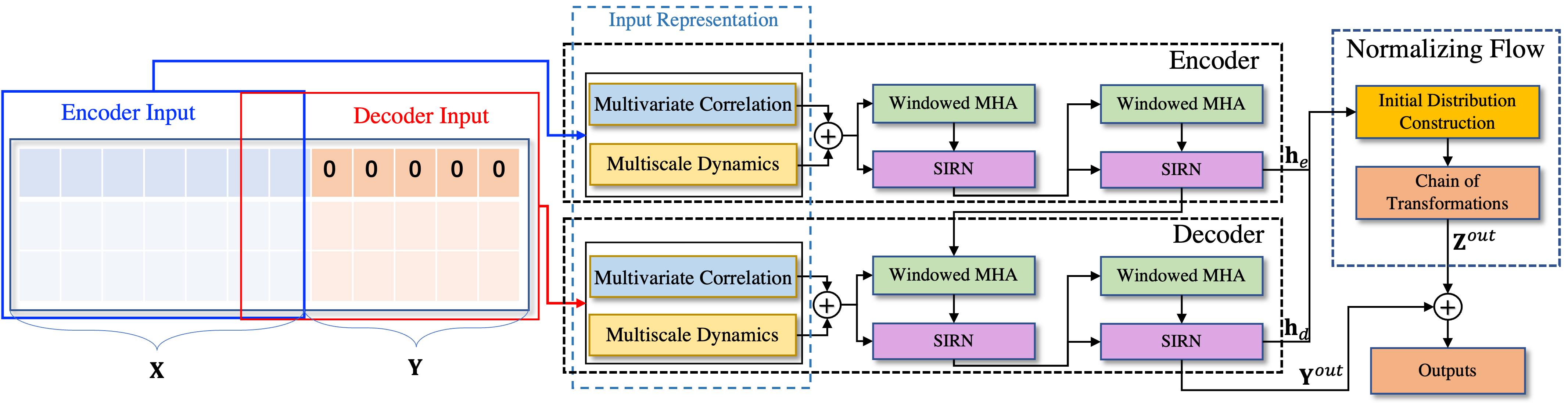}
  \vspace{-1ex}
  \caption{
  The framework overview of \ourmodel.
  In particular, the encoder extracts local patterns with sliding-window multi-head attention (MHA) and explores long-term trends and instant patterns with the proposed SIRN module. 
  The decoder then receives long sequence inputs with the target elements being padded into zeros, measures the weighted composition of multi-faceted temporal patterns, and generates the prediction for target elements. 
  At last, {the normalizing flow block absorbs latent states} yielded in the encoder-decoder architecture and predicts target elements with a chain of invertible transformations directly. 
}
\label{fig:frame}
\vspace{-2.5ex}
\end{figure*}

\section{Related Work}

\subsection{Methods for Time-Series Forecasting}

Many statistical methods have achieved big success in time-series forecasting (TF). 
For instance, ARIMA~\cite{arima} is flexible to subsume multiple types of time-series but the limited scalability strains its further applications. 
Vector Autoregression (VAR)~\cite{var,evar} makes significant progress in multivariate TF by discovering dependencies between high-dimensional variables.
Besides, there exist other traditional methods for the TF problem, such as SVR~\cite{svr}, SVM~\cite{svm}, etc., which also play important roles in different fields.

Another line of studies focuses on deep learning methods for TF, including RNN- and CNN-based models. 
For example, LSTM~\cite{lstm} and GRU~\cite{gru} show their strengths in extracting the long- and short-term dependencies, LSTNet~\cite{lstnet} combines the CNN and RNN to capture temporal dependencies in {the time-series data}, DeepAR \cite{deepar} utilizes the autoregressive {{model}}, as well as the RNN, to model the distribution of future time-series. 
There are also some works focusing on {{CNN models}}~\cite{tcnn,2015time,2018dilated,2021multivariate}, which can capture {{inner patterns}} of the time-series data through convolution.

The Transformer~\cite{attention} has shown its great superiority in NLP problems because of its effective self-attention mechanism, and it has been extended to many different fields successfully. 
There are many attempts to apply the Transformer to {\revise{TF tasks}}, and the main idea lies in aiming to break the bottleneck of efficiency by focusing on the sparsity of the self-attention mechanism. 
The LogSparse Transformer~\cite{logsparse} allows each point to attend to itself and its previous points with exponential step size, Reformer~\cite{reformer} explores the hashing self-attention, Informer~\cite{Informer} utilizes probability estimation to reduce the time and memory complexities, Autoformer~\cite{autoformer} studies the auto-correlation mechanism in place of self-attention.
{All the above models} reduce the complexity of self-attention to $\mathcal{O}(L \log L)$.
The Sparse Transformer~\cite{sparseattn} reduces the complexity to $\mathcal{O}(L \sqrt{L})$ with attention matrix factorization. 
The very recent Longformer~\cite{longformer} and BigBird~\cite{bigbird} adopt a number of attention patterns and can further reduce the complexity to $\mathcal{O}(L)$. 
However, the above reduction of complexity is often at the expense of sacrificing information utilization and the self-attention mechanism might not be reliable when temporal patterns are intricate in the LTTF task.

\subsection{Generative Models}

There are works attempting to learn the distribution of future time-series data.
Gaussian mixture model (GMM)~\cite{gmm} can learn the complex probability distribution with the EM algorithm, but it fails to suit dynamic scenarios.
Wu et al.~\cite{2021dynamic} proposed a generative model for TF by using the dynamic Gaussian mixture.
\cite{pmlr-prob} devises an end-to-end model to make coherent and probabilistic forecasts by generating the distribution of parameters.
In addition, the authors of~\cite{time_autoregress} proposed an autoregressive model to learn the distribution of the data and make the probabilistic prediction.

The variational inference was proposed for generative modeling and  introduced latent variables to explain the observed data~\cite{normalizing}, which provides more flexibility in the inference. 
Both GAN~\cite{gan} and VAE~\cite{vae} show their impressive performances in distribution inference, but the cumbersome training process plus the limited generalization to new data hinder them for wider applications. 
{{Normalizing Flows (NFs) are a family of generative models, an NF is the transformation of a simple distribution that results in a more complex distribution}}. 
{{NF models}} have been applied in many fields successfully to learn intractable distribution, including image generation, noise modeling, video generation, audio generation, etc.
\ourmodel employs the NF as an inner block for LTTF to absorb {{latent states}} in the encoder-decoder architecture, which differentiates itself from prior works.

\section{Problem Statement}
\vspace{-.5ex}

We  introduce the problem definition in this section. 
Given a length-$L$ time-series $ \mathcal{X} = \{ \mathbf{x}_1, \mathbf{x}_2, \cdots \, \mathbf{x}_{L} | \, \mathbf{x}_i \in \mathbb{R}^{d_x} \} $ where $ \mathbf{x}_i $ is not limited to the univariate case (\ie $ d_x \geq 1 $),  the time series forecasting problem takes a length-$L_x$ time-series $\mathbf{X} = \{ \mathbf{x}_{m+1}, \cdots, \mathbf{x}_{m + {L_x}} \} $ as input to predict the future length-$L_y$ time series $ \mathbf{Y} = \{ \mathbf{x}_{n+1},  \cdots, \mathbf{x}_{n + {L_y}} \}$ ($n = m + {L_x}$ and $ m = 1, \cdots, L - L_y $).  
For the sake of clarity, we denote $ \mathbf{Y} = \{ \mathbf{y}_{n+1},  \cdots, \mathbf{y}_{n + {L_y}} \, | \, \mathbf{y}_j \in \mathcal{X} \}$. 
{\em Long-term time-series forecasting} is to predict the future time-series with larger $L_y$.

\section{Methodology}
\vspace{-.5ex}

The framework overview of \ourmodel is shown in~\cref{fig:frame}. 
\ourmodel mainly consists of {\revise{three parts:}} the {\em input representation} block, {\em encoder-decoder} architecture, and {\em normalizing flow} block.
First, the input representation block preprocesses and embeds the input time series accordingly. 
Then, the encoder-decoder architecture explores the local temporal patterns with windowed attention from {\revise{time-series representations}} and examines {\revise{long-term intricate dynamics}} from both stationary and instant perspectives with the help of recurrent network and time-series decomposition. 
Moreover, to improve information utilization, the normalizing flow block leverages {\revise{latent states}} in the recurrent network and generates target series from the latent states directly. 
The technical details of these three components will be introduced in {\revise{the following subsections}}.

\subsection{Input Representation}   \label{sec:input}
\vspace{-1ex}

The time series data exhibits intricate patterns since {\revise{multi-faceted underlying signals}} are often complex and varying.
Given a length-$L$ time series $ \mathcal{X} $, $ \mathcal{X} = \{\mathbf{x}_1, \mathbf{x}_2, \cdots, \mathbf{x}_L | \mathbf{x}_i \in \mathbb{R}^{d_x} \} $ ($d_x \geq 1$), we investigate the underlying multi-faceted relatedness in $\mathcal{X}$ from two perspectives, \ie~{\revise{the}} ``vertical'' feature perspective, and {\revise{the}} ``horizontal'' temporal perspective. 

\begin{figure}[t]
    \begin{subfigure}[t]{0.47\columnwidth}
	\centering
    \includegraphics[width=1.0\textwidth, center]{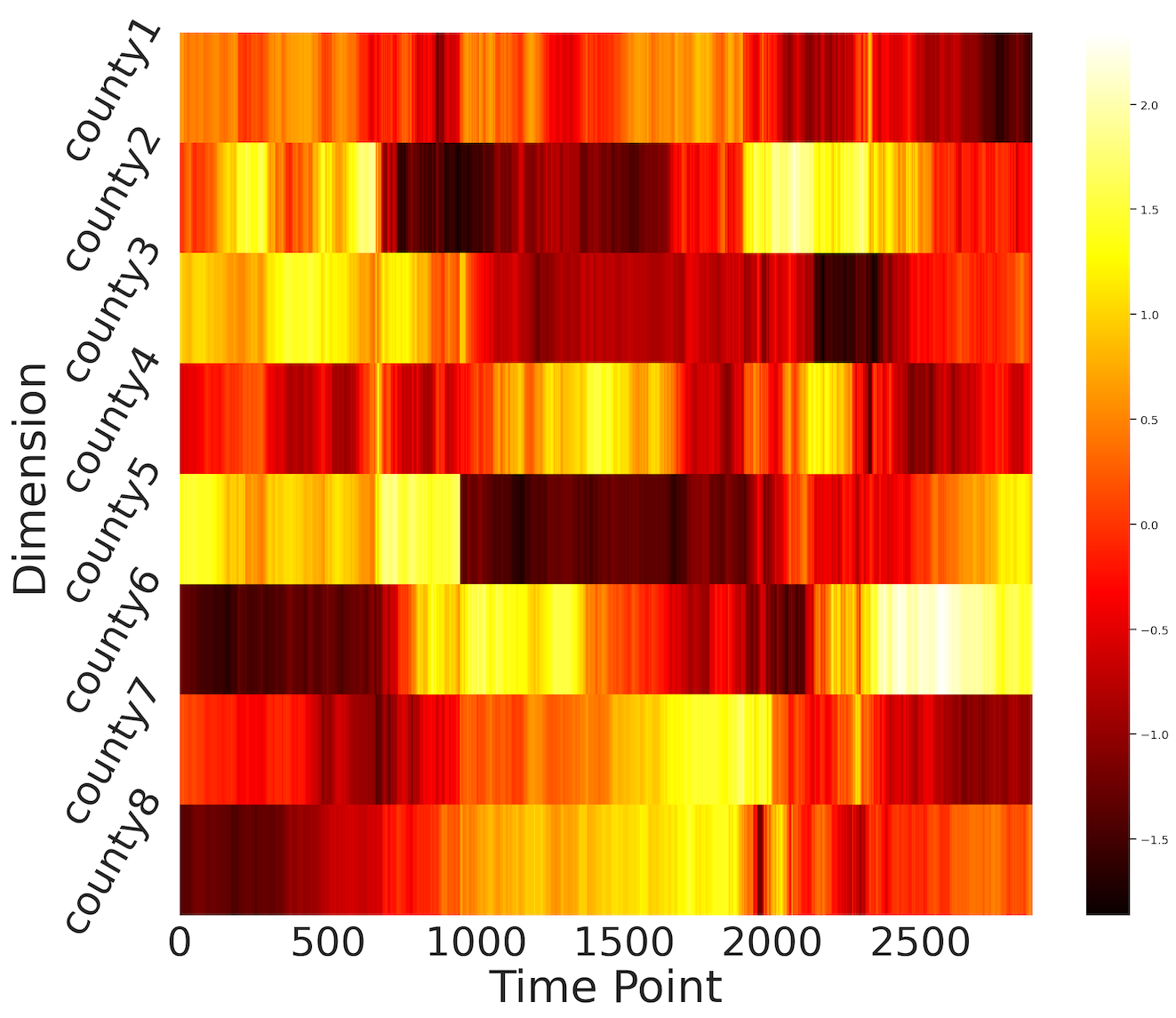}
    \vspace{-3ex}
    \caption{{\revise{Exchange rate.}}}
    \label{fig:multi-var:exchange}
	\end{subfigure}
    \begin{subfigure}[t]{0.49\columnwidth}
	\centering
    \includegraphics[width=1.0\textwidth, center]{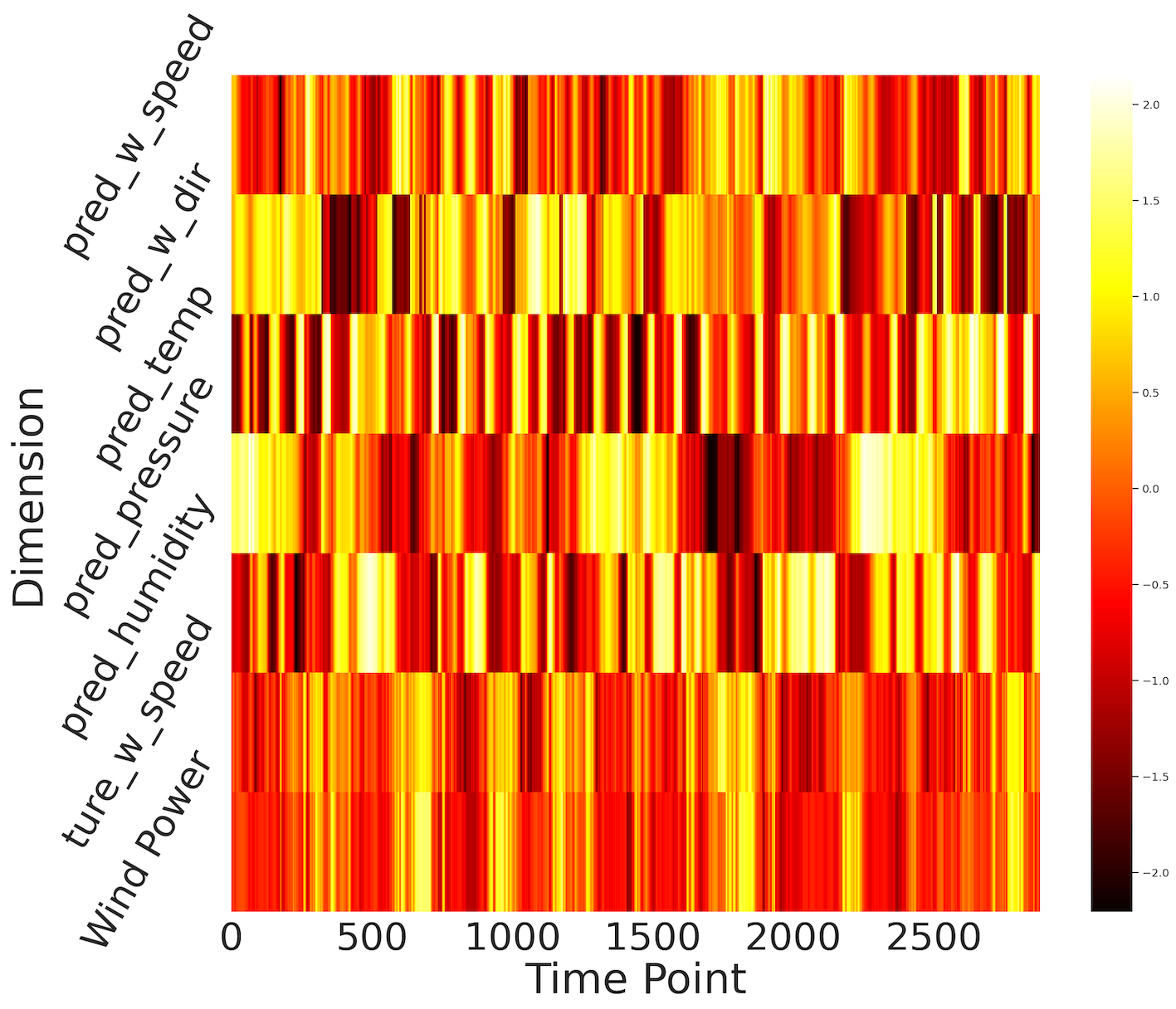}
    \vspace{-3ex}
    \caption{{\revise{Wind power.}}}
    \label{fig:multi-var:wp}
	\end{subfigure}
    \vspace{-1ex}
    \caption{
    {{Different variables of time-series data evolve at varying rhythms and dynamics. The details of these datasets can be found in~\cref{sec:exp:dataset}.}}
    }
    \label{fig:heatmap}
    \vspace{-3ex}
\end{figure}

\subsubsection{\bf Multivariate Correlation}

{{Complex relatedness among different variables in a multivariate time series}} hinders the effectiveness of distinguishing and harnessing important signals for future series prediction. 
On the one hand, the impacts of different variables on forecasting future series differ. 
For instance, the heatmaps in~\cref{fig:heatmap} illustrate {\revise{rhythms}} of different variables in various time-series datasets, it is clear that different variables exhibit distinct relatedness to the target variable, which can also vary over time. 
On the other hand, the well-leveraged dependencies among variables can benefit time-series forecasting.

{\revise{Fast Fourier Transform}} (FFT)~\cite{fft} has been proven to be effective in discovering the correlations for time series data~\cite{fft-t,fft-st, correlation}. 
Inspired by this, we adopt FFT to represent 
{\revise{
implicit multivariate correlations of a length-$L$ time series by exploring the auto-correlation as follows%
}}:
\begin{equation}
    \mathcal{MR}_{\mathcal{XX}} = f^{-1} ( f (\mathcal{X}) f^{\ast} (\mathcal{X}) ) \, ,
\end{equation}

\noindent
where $f$ and $f^{-1}$ denote {\revise{FFT and inverse FFT}}, respectively. 
The asterisk represents a conjugate operation. 
Besides, we employ {\revise{Softmax}} to highlight {\revise{informative variables}} accordingly: 
\begin{equation}    \label{eq:wr}
    \mathbf{W}^{\mathcal{R}} = \text{Softmax} ( \mathcal{MR}_{\mathcal{XX}}) \, .
\end{equation}

\subsubsection{\bf Multiscale Dynamics}

{\revise{Temporal patterns}} are helpful in solving the long-term time-series forecasting problem~\cite{temporal}.
We further examine the temporal patterns by means of multiscale representation. 
Specifically, a time series can present distinct temporal patterns at different temporal resolutions.
In other words, more attention should be paid to {\revise{informative dynamics}} extracted at certain temporal resolutions.

To implement the temporal pattern extraction at different scales, we first devise a temporal resolution set 
$ S \subseteqq \{ second,$ $minute,$ $hour,$ $day,$ $week, month, year\}$ 
for  $\mathcal{X}$. 
Then the sampled time-series set $ {\Gamma}^{S} = \{ {\Gamma}^{S_1}, \cdots, {\Gamma}^{S_K} \} $ is obtained, where $K$ denotes the number of temporal resolutions and $ {\Gamma}^{S_k} $ is the sequence of sampled timestamps at corresponding temporal resolution $ S_k $.  
Afterward, each series in ${\Gamma}^{S}$ is embedded into a latent space with $ d \times L $ dimensionality, such that different series in $ {\Gamma}^{S} $ are additive: 
\begin{equation}    \label{eq:x_s}
\begin{aligned}
    \tilde{{\Gamma}}^{S} 
    & 
    = \mathcal{E}({\Gamma}^{S}) 
    = \{\mathcal{E}({\Gamma}^{S_1}),  \cdots, \mathcal{E}({\Gamma}^{S_K}) \} 
    \\
    & 
    = \{ \tilde{{\Gamma}}^{S_1}, \cdots, \tilde{{\Gamma}}^{S_K} \} \, ,
\end{aligned}
\end{equation}

\noindent
where $\mathcal{E}$ {\revise{denotes an}} embedding operation and $\tilde{{\Gamma}}^{S_k} \in \mathbb{R}^{ d \times L}$ represents the embedded series at a certain temporal resolution $S_k$.
Then the multiscale temporal patterns can be modeled as:
\begin{equation}    \label{eq:H_ts}
\begin{aligned}
    \bar{\Gamma}^{S}
    &
    = \mathbf{W}^{S} \, \text{Concat}(\tilde{{\Gamma}}^{S}) + (\mathbf{b}^{S})^{\prime} 
    \\
    &
    = \sum_{k = 1}^{K} \mathbf{W}_{k}^{S} \, (\tilde{\Gamma}^{S_k})^{\prime} + (\mathbf{b}^{S})^{\prime}
    \, ,
\end{aligned}
\end{equation}
where $\mathbf{W}^{S} \in \mathbb{R}^{ L \times L \times K }$ and $\mathbf{b}^{S} \in \mathbb{R}^{ d \times L}$  are trainable weights and bias, respectively.
The prime symbol denotes the matrix transpose. 
Besides, $ \mathbf{W}_{k}^{S} \in \mathbb{R}^{L \times L} $ denotes the $k$-th sliced matrix of $\mathbf{W}^{S}$.  

\subsubsection{\bf Fusing Multivariate and Temporal Dependencies}

Moreover, to make different variables in multivariate time series more distinguishable w.r.t. their importance for future series, {\revise{we further apply the convolution to take temporal dependencies into account}}, which is defined as follows:
\begin{equation} \label{eq:H_v}
    \mathcal{X}^{v} = \mathbf{W}^{v} \odot (\mathbf{W}^{\mathcal{R}} \, \mathcal{X} 
    + \mathcal{X}) + \mathbf{b}^{v}
    \, ,
\end{equation}
where $\odot$ {{denotes}} the convolution operation, {\revise{and}} $\mathbb{W}^{v} \in \mathbb{R}^{d_x \times d}$ and $\mathbf{b}^{v} \in \mathbb{R}^{d \times L}$ denote weights and bias, respectively. 

Finally, by combining the above multivariate correlations and multiscale dynamics with \cref{eq:wr,eq:H_v}, 
the outcome time-series representation can be obtained as follows:
\begin{equation}    \label{eq:H_in}
    \mathcal{X}^{in} = 
    \mathcal{X}^{v} + \bar{\Gamma}^{S} \, .
\end{equation}

\subsection{Encoder-Decoder Architecture}
\vspace{-1ex}

Our proposed \ourmodel adopts the encoder-decoder architecture for long-term time-series forecasting.

\subsubsection{\bf Attention Mechanism}

The standard attention mechanism~\cite{attention} takes a three-tuple $(query, key, value)$ as input and employs the scaled dot product and Softmax to calculate the weights against the value as: 
$ \text{Attn}(Q, K, V) = \text{Softmax}(\frac{Q K^T}{\sqrt{d_{k}}}) V $, 
where $Q \in \mathbb{R}^{L \times d_k}$, $K \in \mathbb{R}^{L \times d_k}$, and $V \in \mathbb{R}^{L \times d_v}$ represent query, key and value, respectively.

Moreover, the multi-head attention (MHA)~\cite{attention} employs projections for the original query, key, and value  $N$ times, 
and the $i$-th projected query, key, and value can be obtained by $Q_i = Q W^{Q}_{i}$, $ K_i = K W_{i}^{K}$, and $V_i = V W^{V}_{i}$, 
where $W_{i}^{Q} \in \mathbb{R}^{d_k \times d_k / N} $, 
$W_{i}^{K} \in \mathbb{R}^{d_{k} \times d_{K} / N} $,
and $W_{i}^{V} \in \mathbb{R}^{d_v \times d_{v} / N} $. 
{{Afterward, the attention can be applied to these queries, keys, and values in parallel, and the outcome is further concatenated and projected as follows:}} 
%
\begin{equation}    \label{eq:Multi}
\begin{aligned}
    \text{ha}_i = &
                \text{Attn} (Q_i, K_i, V_i), \quad i = 1, 2, \cdots, N   \\
    \text{MHA} & (Q, K, V) = \text{Concat} ( \text{ha}_1, \text{ha}_2, \cdots, \text{ha}_N ) W^o \, .
\end{aligned}
\end{equation}

\textbf{Sliding-Window Attention.}~
{\revise{Duplicated}}~{\revise{messages}} 
exist across different heads in full self-attention \cite{darkattention}. 
{\revise{A time series often shows a strong locality of reference, thus a great deal of information about a point can be derived from its neighbors.
Hence, the full attention message might be too redundant for future series prediction.}}
Given the importance of locality for TF, the sliding-window attention (with fixed window size $w$) allows each point attends to its $\frac{1}{2}w$ neighbors on each side.
Thus, the time complexity of this pattern is $\mathcal{O}(w \times L)$, which scales linearly with input length.
Therefore, we adopt this windowed attention to realize self-attention. 
%

\subsubsection{\bf Stationary and Instant Recurrent Network}

\begin{figure*}[t]
    \centering
    \begin{subfigure}[t]{0.375\textwidth}
	\centering
    \includegraphics[width=0.85\textwidth, center]{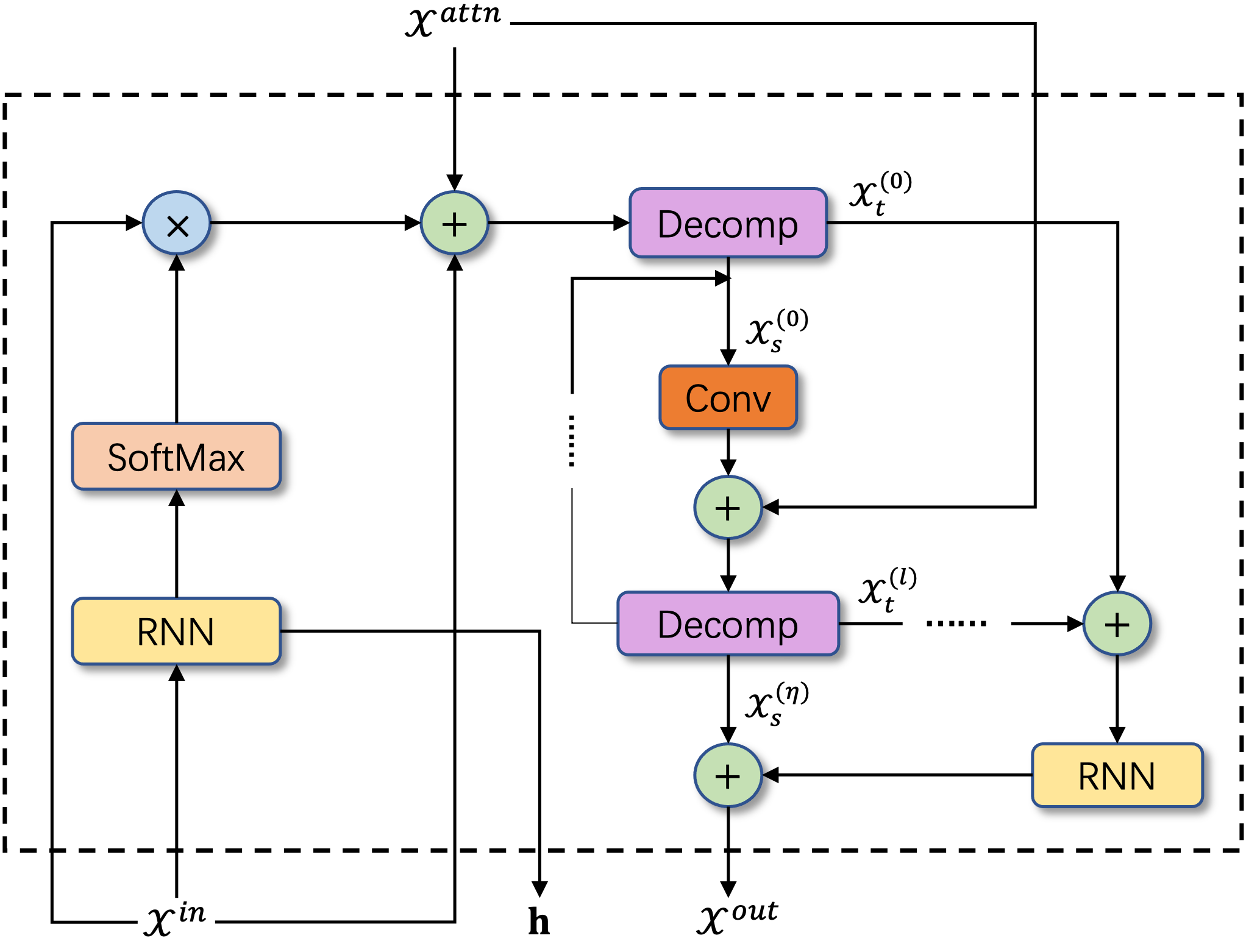}
    \vspace{-3ex}
    \caption{Stationary and instant recurrent network (SIRN).}
    \label{fig:framework-sirn}
	\end{subfigure}
    \begin{subfigure}[t]{0.6\textwidth}
	\centering
    \includegraphics[width=0.95\textwidth, center]{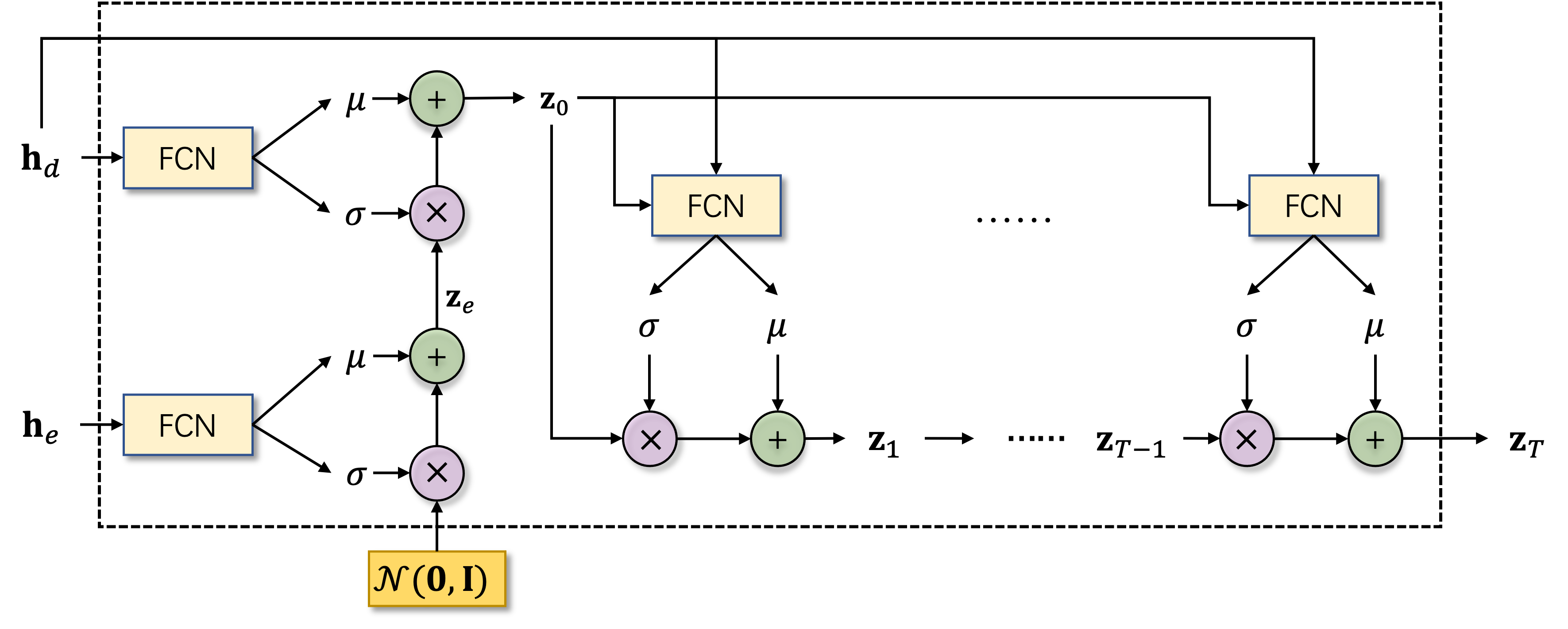}
    \vspace{-3ex}
    \caption{Normalizing flow framework.}
    \label{fig:framework-nf}
	\end{subfigure}
	\vspace{-1ex}
    \caption{
    The architecture of SIRN and the normalizing flow framework.
    (a) 
    The first RNN block embeds the global information of input time-series and the second RNN block represents the aggregated trend information extracted by the decomposition block.
    The decomposition procedure following the initial decomposition can be repeated multiple times. 
    The latent state yielded by the first RNN  will be utilized in the {\em normalizing flow} framework. 
    (b) 
    After initiating the flow of transformations with \cref{z_e,z_0}, the  latent state of decoder is adopted to generate the target variable.  
    }
    \label{fig:lstm_block}
    \vspace{-3ex}
\end{figure*}

Although {\revise{the windowed attention}} can reduce the complexity to $\mathcal{O}(L)$, the information utilization could be sacrificed for LTTF due to {\revise{point-wise sparse connections}}. 
RNNs have achieved big successes in many sequential data applications~\cite{ed-rnn, dual-rnn, lstm-rnn, prediction-rnn} attributed to their capabilities of capturing {\revise{dynamics}} in sequences via cycles in the network of nodes. 
To enhance information utilization without increasing {\revise{time and memory complexities,}} we, therefore, renovate the recurrent network accordingly.
In particular, we not only distill the stationary (trend) and instant (seasonal) temporal patterns from {\revise{input series}} but also integrate the distilled long-term patterns, as well as the aforementioned local temporal patterns, {\revise{into the time-series representation}}. 
The architecture of the proposed Stationary and Instant Recurrent Network (SIRN) is demonstrated 
in \cref{fig:framework-sirn}.

{\revise{Specifically}}, we feed the input representation to the first RNN block (followed by a Softmax) to initialize the global  representation and add it to the local representation, as well as the original input representation, {\revise{as follows}}: 
%
\begin{equation}    \label{eq:X-in-updated}
\begin{aligned}
    \mathcal{X}^{in} 
    =& \text{SoftMax} (\text{RNN}(\mathcal{X}^{in})) \times \mathcal{X}^{in} 
    \\
    &+ \text{MHA}_{W} (\mathcal{X}^{in}) + \mathcal{X}^{in}  \, ,
\end{aligned}
\end{equation}
where  $\text{MHA}_{W} (\cdot)$ denotes the sliding-window attention.
{Note that the RNN block} (followed by Softmax) in the first term of \cref{eq:X-in-updated} aims to capture the global temporal dependency, which can supplement the local dependency captured by the windowed attention.

Though intricate and diverse, the complex temporal patterns in different time-series data can be roughly divided into (coarse-grained) stationary trends and (fine-grained) instant patterns. 
Along this line, we employ the series decomposition introduced in \cite{stl,autoformer} to distill {\revise{stationary and instant}} patterns by capturing {\revise{trend and seasonal}} parts of the time-series data. 
Similar to \cite{autoformer}, we adopt the moving average to capture long-term trends and the residual of {\revise{the original series}} subtracting the moving average as seasonal patterns:
\begin{equation}    \label{eq:x_t}
    \mathcal{X}_t = \text{AvgPool} (\text{Padding} (\mathcal{X}^{in})),  \quad
    \mathcal{X}_s = \mathcal{X}^{in} - \mathcal{X}_t,
\end{equation}
where $ \mathcal{X}_t, \mathcal{X}_s \in \mathbb{R}^{L \times d_x} $ denote the trend and seasonal parts of  $ \mathcal{X}^{in} $, respectively.
{\revise{Then, we use a convolution layer to embed the seasonal pattern.
And, we feed the embedded representation, coupled with the local representation, to another decomposition block for distilling more seasonal patterns.}}
{\revise{This}} distillation process can be implemented in a recurrent way: 

\begin{equation}    \label{eq:recurr-decomp}
\begin{aligned}
    \mathcal{X}_t^{(l)}, \mathcal{X}_s^{(l)} 
    & 
    = \text{Decomp} ( \text{Conv} (\mathcal{X}_s^{(l - 1)}) 
    \\
    &
    + \text{MHA}_{W} (\mathcal{X}^{in} ) ), 
    \,\, l = 1,  \cdots, \eta   
    \, ,
\end{aligned}
\end{equation}

\noindent
where $\text{Decomp}$ denotes~\cref{eq:x_t}, 
$\mathcal{X}_s^{(0)} = \mathcal{X}_s$ and $\mathcal{X}_t^{(0)} = \mathcal{X}_t$.
On the other hand, {\revise{the trend parts}} generated by different decompositions are merged and fed to the second RNN block.
Finally, the distilled multi-faceted temporal dynamics are fused to generate the outcome representation:
\begin{equation}    \label{eq:y_o}
    \mathcal{X}^{out} = \mathbf{W} (\mathcal{X}_s^{(\eta)} + \text{RNN} (\sum_{l = 0}^{\eta} \mathcal{X}_{t}^{(l)})). 
\end{equation}

\subsection{Time Series Prediction with Normalizing Flow}
\vspace{-1ex}

{{The aforementioned SIRN framework adopts RNN to extract global signals.}}
{{In addition, the hidden states yielded by {\revise{RNN}} are beneficial for understanding the distribution of time-series data}}.
Specifically, we design a normalizing-flow block to learn the distribution of hidden states to increase the reliability of {\revise{prediction}}.

\subsubsection{\bf Background of {\em Normalizing Flow}}

A time series 
$\mathcal{X} = \{\mathbf{x}_1,  $  $ \cdots, $  $\mathbf{x}_{L} \}$ can be reconstructed by maximizing the marginal log-likelihood: 
$ \log p(\mathcal{X}) = \sum_{i=1}^{L} \log p(\mathbf{x}_i) $. 
{{Due to}} the intractability of such  log-likelihood, a parametric inference model over the latent variables $\mathbf{z}$, \ie $q(\mathbf{z} | \mathbf{x})$,  was introduced.  
Then,  one can optimize the variational lower bound on the marginal log-likelihood of {\revise{each observation $\mathbf{x}$}} as follows:
\begin{equation}    \label{eq:lpy}
\begin{aligned}
    \log p(\mathbf{x}) \geqslant 
    & 
    \mathbb{E}_{ q(\mathbf{z} | \mathbf{x} )} [\log p(\mathbf{x}, \mathbf{z}) - \log q(\mathbf{z}, \mathbf{x})] 
    \\
    =& 
    \log p(\mathbf{x}) - \mathcal{D}_{KL} (q(\mathbf{z} | \mathbf{x}) \, || \, p(\mathbf{z} | \mathbf{x}))
    \\
    =& \mathcal{L} (\mathbf{x}; \theta) \, ,
\end{aligned}
\end{equation}
where $\mathcal{D}_{KL}(\cdot)$ denotes the Kullback-Leibler divergence. 
When the dimensionality of $\mathbf{z}$ climbs up, the diagonal posterior distribution is often adopted, which is, however, not flexible enough to match the complex true posterior distributions~\cite{2021temporal}. 
To solve this, the Normalizing Flow~\cite{normal} was proposed to build {\revise{flexible posterior distributions}}.

Basically, one can start off with an initial random variable $\mathbf{z}_0$ (with a simple distribution, coupled with a known density function), 
and then apply a chain of invertible transformations $\text{f}_t$, such that the outcome $\mathbf{z}_{T}$ has a more flexible distribution:
\begin{equation}    \label{eq:z_0}
    \mathbf{z}_0 \backsim q ( \mathbf{z}_0 | \mathbf{x} ), \quad 
    \mathbf{z}_t = \text{f}_t ( \mathbf{z}_{t-1}) , \quad  t = 1, \cdots, T \, .
\end{equation}
%
Besides, as long as the Jacobian determinant $ \det \left | \frac{d \mathbf{z}_t}{d \mathbf{z}_{t-1}} \right | $ is available, the transformation can take the following definition:  
\begin{equation}    \label{eq:f_t}
    \text{f}_t (\mathbf{z}_{t-1}) = \mathbf{z}_{t-1} + \mathbf{u} \, g(\mathbf{w}^T \mathbf{z}_{t-1} + {b}) \, ,
\end{equation}

\noindent
where $\mathbf{u}$,  $\mathbf{w}$ and $b$ are parameters,
and $g(\cdot)$ denotes a nonlinear function.

\subsubsection{\bf  Normalizing Flow for LTTF}

The proposed architecture of normalizing flow in \ourmodel is shown in \cref{fig:framework-nf}.

Let $\mathbf{h}$ denote the hidden state yielded by the first RNN block in SIRN.  
Then, draw a random variable from a Gaussian distribution, \ie $\bm{\epsilon} \backsim \mathcal{N} (\mathbf{0}, \mathbf{I})$, 
and the distribution of the hidden state in the encoder can be obtained as: 
\begin{equation}    \label{z_e}
    \mathbf{z}_{e} = \text{FCN}_{\mu}^{(e)} (\mathbf{h}_{e}) + \text{FCN}_{\sigma}^{(e)} (\mathbf{h}_{e}) \cdot {\bm{\epsilon}} \, ,
\end{equation}

\noindent
where $ \text{FCN}_{\mu}^{(e)} $ and $ \text{FCN}_{\sigma}^{(e)} $ are two fully connected networks, 
$ \mathbf{h}_{e} $ denotes the  hidden state in encoder. 
Afterward, we take the latent representation $ \mathbf{z}_{e} $ and the decoder latent state $\mathbf{h}_{d}$ as input to initiate the normalizing flow:
\begin{equation}    \label{z_0}
    \mathbf{z}_0 = \text{FCN}_{\mu}^{(d)} (\mathbf{h}_{d}) + \text{FCN}_{\sigma}^{(d)} (\mathbf{h}_{d}) \cdot \mathbf{z}_{e} \, .
\end{equation}

\noindent
Now that the normalizing flow can be iterated as follows: 
\begin{equation}    \label{z_t}
\begin{aligned}
    \mathbf{z}_t 
    = & \,\,
    \text{FCN}_{\mu}^{(t)} (\mathbf{h}_d, \mathbf{z}_{t-1}) 
    \\
    + & \,\,
    \text{FCN}_{\sigma}^{(t)} (\mathbf{h}_d, \mathbf{z}_{t-1}) \cdot \mathbf{z}_{t-1}, 
    \quad
    t = 1, \cdots, T
    \, .
\end{aligned}
\end{equation}

\noindent
Here, we utilize {\revise{the decoder latent state}} to cascade the message, such that the future series can be generated directly.

\subsection{Loss Function}
\vspace{-1ex}

In order to coordinate with the other parts of \ourmodel, the commonly used log-likelihood is substituted for the MSE (mean squared error) loss function for learning the normalizing flow framework. 
In particular, the random variable sampled from the outcome distribution, \ie  $\mathbf{z}_t$, is deemed as the point estimation of the target series. 
Then, we adopt {\revise{MSE loss functions}} on prediction w.r.t. the target series for both encoder-decoder architecture and normalizing flow framework. 
Finally, the  loss function is defined as follows: 
\begin{equation}    \label{eq:loss}
    \mathcal{L} = 
    \lambda \cdot \text{MSE} (\mathbf{Y}^{out}, \mathbf{Y}) 
    + (1 - \lambda) \cdot \text{MSE} (\mathbf{Z}^{out}, \mathbf{Y})
\end{equation}
%
where $ \mathbf{Y}^{out} $ and $ \mathbf{Z}^{out} $ denote the output of {\revise{decoder and normalizing flow}}, respectively, and $\lambda$ is a trade-off hyperparameter balancing the relative contributions of {\em encoder-decoder} and {\em normalizing flow}.

\section{Experiments}

\subsection{Experiment Settings} 
\vspace{-.5ex}

\subsubsection{\bf Datasets} 
\label{sec:exp:dataset}

We conduct experiments on {\revise{seven}} datasets including five benchmark datasets and {\revise{two}} collected {\revise{datasets}}. 
\cref{tab:dataaset} describes some basic statistics of these datasets.

\textbf{ECL\footnote{https://archive.ics.uci.edu/ml/
datasets/ElectricityLoadDiagrams20112014}}
was collected in {\revise{15-minute intervals}} from 2011 to 2014. 
\revise{%
We select the records from 2012 to 2014 since many zero values exist in 2011~\cite{lstnet}.~%
}%
The processed dataset 
contains the hourly electricity consumption of 321 clients. 
We use 'MT\_321' as the target, and the train/val/test is 12/2/2 months.

\textbf{Weather\footnote{https://www.bgc-jena.mpg.de/wetter/}}
was recorded in {\revise{10-minute}} intervals from 07/2020 to 07/2021.  
{\revise{There exist 21 meteorological indicators}}, 
\eg~\revise{the amount of rain}, \revise{humidity}, etc. 
We choose temperature as the target, and the train/val/test is 10/1/1 months.

\textbf{Exchange}~\cite{lstnet}
records the daily exchange rates of eight countries 
from 1990 to 2016. 
We use the exchange rates of Singapore as the target, 
The train/val/test is 16/2/2 years. 

\textbf{ETT}~\cite{Informer}
{\revise{records the electricity transformer temperature.}}
{\revise{Every}} data point consists of six power load features and the target value is ``oil temperture''.
This dataset is separated into \{\textit{ETTh1, ETTh2}\}  and 
\{\textit{ETTm1, ETTm2}\} for 1-hour-level and 15-minute-level observations, respectively. 
We use \textit{ETTh1} and \textit{ETTm1} as our datasets. 
The train/val/test are 12/2/2 and 12/1/1 months for \textit{ETTh1} and  \textit{ETTm1}, respectively. 

\textbf{Wind} (Wind Power)\footnote{We collect this dataset and publish it at \url{https://github.com/PaddlePaddle/PaddleSpatial/tree/main/paddlespatial/datasets/WindPower}.}
records the generated wind power of a wind farm {\revise{in 15-minute intervals}} from 01/2020 to 07/2021. 
The train/val/test is 12/1/1 months. 

\revise{%
\textbf{AirDelay}
was collected from the ``On-Time'' database in the TranStas data library\footnote{%
https://www.transtats.bts.gov. The processed dataset is available at \url{https://github.com/PaddlePaddle/PaddleSpatial/tree/main/paddlespatial/datasets/AirDelay}.}.~%
We extracted the flights arrived at the airports in Texas and examined arrival delays in the first month of the year 2022,  
and the canceled flights were removed.
Note that the time interval of this dataset is varying.  
This dataset was split into train/val/test as 7:1:2.%
}

\begin{table}[t]
    \centering
    \setlength\tabcolsep{1.0pt}
    \renewcommand{\arraystretch}{1.05}
    \caption{
    Statistical descriptions of the time-series datasets.
    }   
    \label{tab:dataaset}
    \vspace{-1ex}
    \begin{tabular}{c|c|c|c|c|c}
    \toprule
    {Datasets} & 
    {\# Dims.} & 
    {Time Span} & 
    {\# Points} &
    {Target Variable} & 
    {{\revise{Interval}}}
    \\  
    \midrule 
    ECL & 321 & 01/2012 - 12/2014 & 26304 & MT\_321 & 1 hour \\
    Weather & 21 & 01/2020 - 06/2021 &36761 & Temperature & 10 mins \\
    Exchange & 8 & 01/1990 - 12/2016 & 7588 & Country8 & 1 day \\
    ETTh1 & 7 & 07/2016 - 07/2018 & 17420 & OT & 1 hour \\
    ETTm1 & 7 & 07/2016 - 07/2018 &69680 & OT & 15 mins \\
    Wind & 7 & 01/2020 - 05/2021 & 45550 & Wind\_Power & 15 mins \\
    AirDelay & 6 & 01/01 - 01/31, 2022 & 54451 & ArrDelay & -- \\
    \bottomrule
    \end{tabular}
    \vspace{-2.5ex}
\end{table}


\subsubsection{\bf Baselines}

We compare \ourmodel with 9 {\revise{baselines}}, \ie 5 {\revise{Transformer methods}} (Autoformer, Informer, Reformer, Longformer, and LogTrans), 2 {\revise{RNN methods}} (GRU and LSTNet), and 2 other deep methods (TS2Vec and N-Beats).

\begin{table*}[tbp]
    \centering
    \small
    \tabcolsep 0.03in
    \caption{
    {\revise{Comparisons of multivariate LTTF results (the best and 2nd best scores are boldfaced and underlined, resp.).}} 
    }
    \vspace{-1.5ex}
    \begin{tabular}{c|c|cc|cc|cc|cc|cc|cc|cc|cc}
         \hline
         \hline
         \multicolumn{2}{c}{\multirow{2}{*}{Model}} & \multicolumn{10}{c|}{{{Transformer-based}}} & \multicolumn{4}{c|}{{RNN-based}} & \multicolumn{2}{c}{{Others}} \\
         \cline{3-18}
         \multicolumn{2}{c|}{~}& 
         \multicolumn{2}{c|}{\ourmodel} & 
         \multicolumn{2}{c|}{Longformer \cite{longformer}} &
         \multicolumn{2}{c|}{Autoformer \cite{autoformer}} &
         \multicolumn{2}{c|}{Informer \cite{Informer}} & 
         \multicolumn{2}{c|}{Reformer \cite{reformer}} &
         \multicolumn{2}{c|}{LSTNet \cite{lstnet}} & 
         \multicolumn{2}{c|}{GRU \cite{gru}} &
         \multicolumn{2}{c}{N-beats\cite{nbeats}} \\
         \hline
         \multicolumn{2}{c|}{Metric}&MSE&MAE&MSE&MAE&MSE&MAE&MSE&MAE&MSE&MAE&MSE&MAE&MSE&MAE&MSE&MAE\\
        \hline
        \multirow{4}{*}{\rotatebox{90}{ECL}}
         &96&\underline{0.2124}&\underline{0.3193}
         &0.3156&0.3939&\textbf{0.2018}&\textbf{0.3100}&0.5423&0.5568&0.9865&0.7795&1.1002&0.8066&0.7292&0.6274& 1.3759& 0.8753  \\
         &192&\textbf{0.2378}&\textbf{0.3456}
         &\underline{0.3371}&\underline{0.4169}&0.3579&0.4277&0.5304&0.5549&1.0119&0.7831&1.0965&0.8048&1.0093&0.7679&1.3228& 0.8660\\
         &384&\textbf{0.2643}&\textbf{0.3620}
         &\underline{0.3976}&\underline{0.4183}&0.4670&0.5019&0.6429&0.5921&1.0883&0.7867&1.1034&0.8057&1.0548&0.7898& 1.3911& 0.8709  \\
         &768&\textbf{0.3396}&\textbf{0.4092}
         &0.5651&\underline{0.5182}&\underline{0.5525}&0.5598&0.9534&0.7789&1.0624&0.7913&1.1132&0.8085&1.0651&0.7938 & 1.3645&0.8585 \\
         \hline
         \multirow{4}{*}{\rotatebox{90}{Weather}}
         &48&\textbf{0.3216}&\underline{0.3433}
         &\underline{0.3475}&0.3630&0.4552&0.4340&0.3929&\textbf{0.3231}&0.5099&0.4506&0.7852&0.6769&0.6569&0.5438& 0.6171& 0.5346  \\
         &192&
         \textbf{0.4129}&\textbf{0.4170}
         &\underline{0.4259}&\underline{0.4235}&0.4965&0.4711&0.4396&0.4332&0.6960&0.5852&0.7858&0.6770&0.7548&0.6025 & 0.6121 & 0.5402\\
         &384&
         \textbf{0.4997}&\textbf{0.4847}
         &\underline{0.5518}&\underline{0.5030}&0.5832&0.5255&0.5848&0.5197&0.7525&0.6231&0.8063&0.6886&0.7679&0.6070 & 0.6032 & 0.5067\\
         &768&
         \textbf{0.6146}&\textbf{0.5603}
         &0.6734&0.5769&\underline{0.6429}&\underline{0.5682}&0.7051&0.5881&0.7883&0.6529&0.8303&0.7003&0.7671&0.6100 & 0.5944& 0.5143 \\
         \hline
         
        \multirow{4}{*}{\rotatebox{90}{Exchange}}
         &48&
         \textbf{0.0764}&\textbf{0.2093}
         &0.1736&0.3314&\underline{0.1431}&\underline{0.2892}&0.2310&0.3841&0.3653&0.4952&1.0319&0.8623&1.1399&0.9119 & 2.1053 & 1.0750\\
         &96&
         \textbf{0.1193}&\textbf{0.2607}
         &0.3519&0.4829&\underline{0.2021}&\underline{0.3586}&0.3079&0.4488&0.9120&0.7731&1.0260&0.8648&1.3953&0.9837 & 1.8161 & 0.9896\\
         &192&
         \textbf{0.2900}&\textbf{0.4187}
         &0.6145&0.6393&\underline{0.4249}&\underline{0.5486}&0.5902&0.6306&1.1195&0.8713&0.9954&0.8562&1.3754&0.9800 & 1.8113& 0.9899 \\
         &384&
         \textbf{0.4730}&\textbf{0.5369}
         &\underline{0.8105}&\underline{0.7513}&1.2798&0.9983&0.8630&0.7953&1.2748&0.9435&0.9642&0.8457&1.3801&0.9858 & 2.4088 & 1.1708\\
         \hline
         
        \multirow{4}{*}{\rotatebox{90}{ETTm1}}
         &96&
         \textbf{0.6854}&\textbf{0.5901}
         &1.0947&0.7079&\underline{0.8586}&\underline{0.6591}&1.0921&0.7023&1.6397&0.9771&1.6250&0.9045&1.7469&0.9714& 1.2350 & 2.3957 \\
         &192&
         \textbf{0.7856}&\textbf{0.6387}
         &1.2555&0.7644&\underline{0.9406}&\underline{0.6958}&1.2657&0.7898&1.6499&0.9659&1.6012&0.9080&1.7223&0.9592&1.2253 & 2.3467 \\
         &384&
         \textbf{0.9298}&\textbf{0.6988}
         &1.2303&0.7786&\underline{1.1112}&\underline{0.7593}&1.3849&0.8459&1.6396&0.9783&1.5063&0.8916&1.5815&0.9124&1.2149 & 2.2922 \\
         &768&
         \textbf{0.9835}&\textbf{0.7193}
         &\underline{1.2247}&\underline{0.7816}&1.2974&0.7940&1.3537&0.8492&1.6121&0.9501&1.3637&0.8604&1.3437&0.8425&1.2420 & 2.3629 \\
         \hline
         
        \multirow{4}{*}{\rotatebox{90}{ETTh1}}
         &96&
         \textbf{0.6978}&\textbf{0.5623}
         &\underline{0.7276}&0.5894&0.7515&\underline{0.5727}&0.8901&0.6498&0.9794&0.6926&1.1763&0.7884&1.0490&0.7312& 1.6426 & 0.9352 \\
         &192&
         \textbf{0.8444}&\textbf{0.6249}
         &\underline{0.9074}&0.6566&0.9346&\underline{0.6375}&1.0463&0.7082&1.0157&0.7156&1.1850&0.7870&1.0850&0.7469 & 1.7524 & 0.9709\\
         &384&
         \textbf{0.9708}&\textbf{0.6767}
         &\underline{0.9804}&\underline{0.6919}&1.1832&0.7347&1.1237&0.7383&1.0673&0.7298&1.2493&0.8082&1.1081&0.7471 & 1.7703& 0.9778 \\
         &768&\underline{1.0827}&\underline{0.7275}
         &\textbf{1.0501}&\textbf{0.7083}&1.2562&0.7676&1.1047&0.7292&1.1105&0.7447&1.5301&0.9207&1.1275&0.7524 & 1.7656&0.9822 \\
         \hline
         
        \multirow{5}{*}{\rotatebox{90}{Wind}}
         &48&
         \textbf{0.9479}&\textbf{0.6539}
         &\underline{0.9605}&0.6767&1.3522&0.8099&1.0056&\underline{0.6552}&1.1881&0.7949&1.3874&0.9246&1.1599&0.8022& 1.5667 & 0.8727\\
         &96&
         \textbf{1.1725}&\textbf{0.7641}
         &1.2467&\underline{0.7698}&1.4859&0.8702&\underline{1.2371}&0.7994&1.3283&0.8529&1.4489&0.9441&1.2797&0.8455& 1.6842&0.9074 \\
         &192&
         \textbf{1.3291}&\textbf{0.8464}
         &1.4829&\underline{0.8487}&1.6118&0.9172&1.5022&0.8489&1.4074&0.8980&1.4794&0.9508&\underline{1.3779}&0.8866 & 1.6146& 0.8839 \\
         &384&
         \textbf{1.3644}&\textbf{0.8692}
         &1.5479&0.8830&1.7363&0.9585&1.5002&\underline{0.8747}&1.4541&0.9190&1.4966&0.9541&\underline{1.3818}&0.8897 & 1.5746& 0.8551 \\
         &768&
         \textbf{1.3698}&\textbf{0.8905}
         &1.4995&\underline{0.8954}&1.6629&0.9426&1.5152&0.8956&1.5215&0.9540&1.4813&0.9471&\underline{1.4580}&0.9253& 1.6176 & 0.9009 \\
         \hline

         \multirow{4}{*}{\rotatebox{90}{AirDelay}}
         & 96 &\textbf{0.7491} & \textbf{0.5702}&0.7746 &0.5984 &0.7959 &0.6041&0.7663&0.5904&0.7719&0.5961&0.7781&0.6058&\underline{0.7675}&\underline{0.5859}&0.7961&0.5940\\
         & 192 & \textbf{0.7523}& \textbf{0.5689} &0.7770 &0.5980 &0.7865&0.5934&\underline{0.7659}&\underline{0.5876}&0.7724&0.5960&0.7782&0.6053&0.7705&0.5903&0.8041&0.5961\\
         & 384 &\textbf{0.7560} & \textbf{0.5718}& 0.7876&0.6051 &0.7896&0.5889&\underline{0.7729}&\underline{0.5841}&0.7735&0.5963&0.7784&0.6049&0.7739&0.5968&0.8082&0.5988\\
         & 768 & \textbf{0.7614}& \textbf{0.5729} &0.8081&0.6209&0.7952&\underline{0.5921}&0.7868&0.6061&\underline{0.7749}&0.5956&0.7797&0.6044&0.7750&0.5961&0.8099&0.5981\\
         \hline
         \hline
    \end{tabular}
    \label{tab:multi}
    \vspace{-2.5ex}
\end{table*}

\begin{enumerate}[leftmargin=0.5cm,label=$\bullet$]
    
    \item {\bf GRU}~\cite{gru}: 
    {\revise{%
    GRU employs the gating mechanism such that each recurrent unit adaptively captures temporal signals in the series.
    In this work, we adopt a 2-layer GRU.
    }}
    
    \item {\bf LSTNet}~\cite{lstnet}:
    {\revise{%
    LSTNet combines the convolution and recurrent networks to extract short-term dependencies among variables and long-term trends in the time series.
    Note that, to simplify the parameter tuning, the highway and skip connection mechanisms are omitted.
    }}
    
    \item {\bf N-Beats}~\cite{nbeats}: 
    {\revise{N-Beats was proposed to address time-series forecasting via a deep model on top of the backward and forward residual links and a very deep stack of fully-connected layers.}}~%
    We implement N-Beats for multivariate LTTF with {\revise{suggested settings}}. 
    
    \item {\bf Reformer}~\cite{reformer}: 
    Reformer uses locality-sensitive hashing (LSH) attention and reversible residual layers to reduce the computation complexity. 
    We implement Reformer by setting the bucket\_length and the number of rounds for LSH attention as 24 and 4, respectively.
    
    \item {\bf Longformer}~\cite{longformer}: 
    {\revise{Longformer combines the windowed attention with a task motivated global attention to scale up linearly as the sequence length grows.}}
    
    \item {\bf LogTrans}~\cite{logsparse}: 
    LogTrans breaks the memory bottleneck of Transformer for LTTF via 
    producing queries and keys with the help of causal convolutional self-attention. 
    The number of the LogTransformer blocks is set to 2 and the sub\_len of the sparse-attention is set to 1.
    
    \item {\bf Informer}~\cite{Informer}: 
    Informer {\revise{proposes the ProbSparse slef-attention}} to reduce {\revise{time and memory complexities}}, and
    {\revise{handles the long-term sequence}} with {\revise{self-attention distilling operation and generative style decoder}}.

    \item {\bf Autoformer}~\cite{autoformer}: 
    Autoformer renovates the series decomposition with the help of auto-correlation mechanism, and put the series decomposition as a basic inner block of the deep model. 
    
    \item {\bf TS2Vec}~\cite{ts2vec}: 
    TS2Vec is a universal framework for learning representations of time series. It performs contrastive learning in a hierarchical way over augmented context views, which leads to the robust contextual representation for each timestamp. 
    We implement TS2Vec for univariate LTTF with the suggested settings. 
    
\end{enumerate}

All baselines employ the one-step prediction strategy. 
For the RNN-based methods, the number of hidden states is chosen from  $\{16, 24, 32, 64\}$. 
{\revise{For}} the Transformer-based methods, 
the number of heads of the self-attention is 8 and the dimensionality is set as 512 for all attention mechanisms in the experiments.
Moreover, the sampling factor of the self-attention is set to 1 for both Informer and Autoformer,  other settings are the same as suggested by~\cite{autoformer}. 
{\revise{All}} Transformer-based baselines (except Autoformer) use the same embedding {\revise{method}} applied to the Informer. 
As suggested by~\cite{autoformer}, we omit the position embedding and keep the value embedding and timestamp embedding for Autoformer.


\subsubsection{\bf Implementation Details}

\ourmodel{\footnote{The source code of \ourmodel~is available at \url{https://github.com/PaddlePaddle/PaddleSpatial/tree/main/research/Conformer}.}} includes a 2-layer encoder and a 1-layer decoder, as well as a 2-layer normalizing flow block. 
The window size of the sliding-window attention is 2, and $\lambda$ in \cref{eq:loss} is set to $ 0.8 $. 
We use {\revise{an}} Adam optimizer, and the initial learning rate is $1\times 10^{-4}$.
The batch-size is 32 and the training process employs early stopping within 10 epochs.
In addition, 
we use MAE (mean absolute error) and MSE (mean squared error) as the evaluation metrics. 

An input-$L_x$-predict-$L_y$ window is applied to roll the train, validation and test sets with stride one time step, respectively.
{\revise{This}} setting is adopted for all datasets.
The input length $L_{x}$ is 96 and the predict length $L_{y}$ is chosen from \{48, 96, 192, 384, 768\} on all datasets. 
%
The averaged  results in 5 runs are reported.
{\revise{All models}} are implemented in PyTorch and trained/tested on a Linux machine with one A100 40GB GPU.

%
{\revise{All of}} the RNN blocks in \ourmodel are implemented with GRU. 
Under the multivariate LTTF {\revise{setting}}, we adopt 1-layer GRU and 2-layer GRU for encoder and decoder, respectively. 
Under the univariate LTTF {\revise{setting}}, both the encoder and decoder adopt 1-layer GRU.

\begin{table*}[t]
    \centering
    \small
    \caption{\revise{
    Multivariate LTTF with time-determined lengths (boldface and underline for the best and 2nd best scores).  
    }}
    \tabcolsep 0.02in
    \vspace{-1.5ex}
    {
    \begin{tabular}{c|c|cc|cc|cc|cc|cc|cc|cc|cc}
         \hline
         \hline
         \multicolumn{2}{c|}{\multirow{2}{*}{Model}} & \multicolumn{10}{c|}{{\revise{Transformer-based}}} & \multicolumn{4}{c|}{\revise{RNN-based}} & \multicolumn{2}{c}{\revise{Others}} \\
         \cline{3-18}
         \multicolumn{2}{c|}{~} & 
         \multicolumn{2}{c|}{\ourmodel} & 
         \multicolumn{2}{c|}{Longformer \cite{longformer}} &
         \multicolumn{2}{c|}{Autoformer \cite{autoformer}} &
         \multicolumn{2}{c|}{Informer \cite{Informer}} & 
         \multicolumn{2}{c|}{Reformer \cite{reformer}} &
         \multicolumn{2}{c|}{LSTNet \cite{lstnet}} & 
         \multicolumn{2}{c|}{GRU \cite{gru}} &
         \multicolumn{2}{c}{N-beats\cite{nbeats}} \\
         \hline
         \multicolumn{2}{c|}{Metric}&MSE&MAE&MSE&MAE&MSE&MAE&MSE&MAE&MSE&MAE&MSE&MAE&MSE&MAE&MSE&MAE\\
         \hline
         \multirow{4}{*}{\rotatebox{90}{ETTh1}}
         &1D&0.4158 &0.4434 &\textbf{0.3924} & \textbf{0.4316}& 0.4248 & 0.4374 &  \underline{0.3985} & \underline{0.4345} & 0.6987 & 0.5640 & 1.1549 & 0.7713 & 0.9265 & 0.6771 & 0.9027 & 1.5612 \\
         ~&1W& \textbf{0.7326}& \textbf{0.5785}&\underline{0.7583} & \underline{0.5997} & 0.8682 & 0.6130 & 0.8585 & 0.6362 & 1.0286 & 0.7207 & 1.2254 & 0.8020 & 1.1184 & 0.7571 & 0.9388 & 1.6576\\
         ~&2W&\textbf{ 0.8661}& \textbf{0.6400}& \underline{0.9328 }& \underline{0.6767} & 1.1191 & 0.7126 & 1.0686 & 0.7054 & 1.0796 & 0.7378 & 1.2231 & 0.7964 & 1.1095 & 0.7456 & 0.9533 & 1.6852 \\
         ~&1M&\underline{0.9845} & \underline{0.6887}&\textbf{ 0.9205 }&\textbf{ 0.6763} & 1.2151 & 0.7567 & 1.0292 & 0.6887 & 1.1092 & 0.7451 &1.6012 & 0.9525 & 1.1341 & 0.7566 & 1.0084 & 1.8501\\
    \hline
        \multirow{3}{*}{\rotatebox{90}{ETTm1}}
         &1D& \textbf{0.6854}&\textbf{0.5901}
         &1.0947&0.7079&\underline{0.8586}&\underline{0.6591}&1.0921&0.7023&1.6397&0.9771&1.6250&0.9045&1.7469&0.9714& 1.2350 & 2.3957 \\
         ~&1W& \textbf{0.9540} & \textbf{0.7009} & 1.2416 & 0.7912 &\underline{1.2016}& \underline{0.7660} & 1.4284 & 0.8551 & 1.4008 & 0.8697 & 1.3692 & 0.8576 & 1.3852 & 0.8551 & 1.2322&2.3267 \\
         ~&2W&\textbf{1.0948} &\textbf{0.7583} & 1.2463 & \underline{0.7871} &1.5101 & 0.8469 & 1.2857 & 0.8177 & \underline{1.2233} & 0.7952 & 1.2931 & 0.8326 & 1.2245 & 0.7945&1.2540&2.3790 \\
    \hline
    \hline
    \end{tabular}}
    \label{tab:time_inv}
    \vspace{-1ex}
\end{table*}

\begin{table*}[tbp]
    \centering
    \small
    \tabcolsep 0.03in
    \caption{
    {\revise{Comparisons of univariate LTTF results (the best and 2nd best scores are boldfaced and underlined, resp.).}}
    }
    \vspace{-1.5ex}
    \begin{tabular}{c|c|cc|cc|cc|cc|cc|cc|cc|cc}
         \hline
         \hline
         \multicolumn{2}{c}{\multirow{2}{*}{Model}} & \multicolumn{10}{c|}{\revise{Transformer-based}} & \multicolumn{4}{c|}{\revise{RNN-based}} & \multicolumn{2}{c}{\revise{Others}} \\
         \cline{3-18}
         \multicolumn{2}{c|}{~}&
         \multicolumn{2}{c|}{\ourmodel} & 
         \multicolumn{2}{c|}{Autoformer \cite{autoformer}} &
         \multicolumn{2}{c|}{Informer \cite{Informer}} &
         \multicolumn{2}{c|}{Reformer \cite{reformer}} &
         \multicolumn{2}{c|}{LogTrans \cite{logsparse}} &
         \multicolumn{2}{c|}{LSTNet \cite{lstnet}} &
         \multicolumn{2}{c|}{GRU \cite{gru}} &
         \multicolumn{2}{|c}{TS2VEC\cite{ts2vec}} \\
         \hline
         \multicolumn{2}{c|}{Metric}&MSE&MAE&MSE&MAE&MSE&MAE&MSE&MAE&MSE&MAE&MSE&MAE&MSE&MAE&MSE&MAE\\
         \hline
         
         \multirow{4}{*}{\rotatebox{90}{ECL}}
         &96&
         \underline{0.3481}&\underline{0.4587}
         &0.4457&0.5295&\textbf{0.3182}&\textbf{0.4425}&0.7945&0.7286&0.6378&0.6859&0.9617&0.7872&0.8348&0.7358& 1.1987 & 2.1247\\
         &192&
         \textbf{0.3565}&\textbf{0.4629}
         &0.5327&0.5787&\underline{0.3906}&\underline{0.4968}&0.8575&0.7465&0.6373&0.6837&0.9929&0.7904&0.9379&0.7700& 1.1557 & 2.0041 \\
         &384&
         \textbf{0.3659}&\textbf{0.4656}
         &0.6592&0.6506&\underline{0.4352}&\underline{0.5260}&0.9269&0.7666&0.6369&0.6807&1.0074&0.7928& 0.9685 & 0.7784 & 1.1863 & 2.0864 \\
         &768&
         \textbf{0.4283}&\textbf{0.4989}
         &0.7821&0.7190&\underline{0.4987}&\underline{0.5674}&0.9521&0.7711&0.6468&0.6815&1.0429&0.8034&0.9917&0.7847& 1.0891 & 1.7940 \\
         \hline
         
         \multirow{4}{*}{\rotatebox{90}{Weather}}
         &48&
         \textbf{0.0846}&\textbf{0.2025}
         &0.1719&0.3073&\underline{0.0914}&\underline{0.2183}&0.1809&0.3202&0.3231&0.4460&0.4818&0.5985&0.3660&0.4689& 1.6989 & 3.8099 \\
         &192&
         \textbf{0.1898}&\textbf{0.3114}
         &0.2365&0.3562&\underline{0.2008}&\underline{0.3330}&0.3597&0.4651&0.3115&0.4270&0.4947&0.5994&0.4505&0.5232& 1.6884 & 3.7227 \\
         &384&
         \textbf{0.3006}&\textbf{0.4041}
         &0.3249&0.4345&\underline{0.3095}&\underline{0.4267}&0.4386&0.5198&0.3500&0.4554&0.5218&0.6171&0.4869&0.5383& 1.4820 & 3.1272 \\
         &768&
         \textbf{0.4005}&\textbf{0.4811}
         &0.4665&0.5314&0.5536&0.5891&0.4493&0.5240&\underline{0.4297}&\underline{0.5065}&0.5414&0.6302&0.5397&0.5508&1.7431 & 3.9300 \\
         \hline
         
         \multirow{4}{*}{\rotatebox{90}{Exchange}}
         &48&
         \textbf{0.0676}&\textbf{0.2059}
         &\underline{0.1446}&\underline{0.2892}&0.2308&0.3840&0.3655&0.4954&0.2768&0.4338&1.0319&0.8623&1.1399&0.9858& 1.2644 & 2.2181 \\
         &96&
         \textbf{0.1168}&\textbf{0.2693}
         &\underline{0.1369}&\underline{0.2935}&0.1827&0.3408&1.5431&1.0742&0.1976&0.3432&1.7163&1.2134&2.2484&1.3663&1.2797 & 2.2738 \\
         &192&
         \textbf{0.2107}&\textbf{0.3751}
         &0.4256&0.5549&\underline{0.3889}&\underline{0.5021}&2.0076&1.2959&0.5285&0.6217&1.6056&1.1823&2.5891&1.5426&1.2846 & 2.3074\\
         &384&
         \textbf{0.4591}&\textbf{0.5770}
         &1.2899&1.0128&1.1126&0.7888&2.2899&1.4411&\underline{0.5520}&\underline{0.6415}&1.5664&1.1789&2.5353&1.5355&1.2947 & 2.3176 \\
         \hline
         
         \multirow{4}{*}{\rotatebox{90}{ETTm1}}
         &96&
         \textbf{0.0655}&\textbf{0.1827}
         &\underline{0.0733}&0.1979&0.0793&\underline{0.1945}&0.1481&0.2865&0.0752&0.2008&0.1583&0.2952&0.3128&0.4853& 1.0667 & 1.7871 \\
         &192&
         \textbf{0.0898}&\textbf{0.2237}
         &0.1018&0.2445&0.1124&0.2407&0.2135&0.3546&\underline{0.0906}&\underline{0.2277}&0.2099&0.3424&0.3007&0.4511& 0.9113 & 1.4667 \\
         &384&\underline{0.1032}&\underline{0.2549}
         &0.1175&0.2746&0.2643&0.4057&0.2699&0.4092&0.1039&0.2560&\textbf{0.0980}&\textbf{0.2479}&0.2569&0.3891& 1.0060 & 1.7039 \\
         &768&\textbf{0.1194}&\underline{0.2770}
         &0.2058&0.3496&0.4202&0.5488&0.2017&0.3470&0.1219&\textbf{0.2693}&\underline{0.1197}&0.2778&0.1693&0.3138& 0.9859 & 1.6759 \\
         \hline
         
         \multirow{4}{*}{\rotatebox{90}{ETTh1}}
         &96&
         \textbf{0.1139}&\textbf{0.2717}
         &0.1484&0.3163&0.1517&0.3164&0.3425&0.4650&\underline{0.1362}&\underline{0.3040}&0.6936&0.7430&0.4917&0.6045& 1.5564 & 3.2653 \\
         &192&
         \underline{0.1452}&0.3114
         &0.1456&\textbf{0.3095}&0.1581&0.3264&0.4233&0.5264&\textbf{0.1435}&\underline{0.3108}&0.8762&0.8584&0.4501&0.5678& 1.5088 & 3.0289 \\
         &384&
         \textbf{0.1431}&\textbf{0.3071}
         &\underline{0.1478}&\underline{0.3087}&0.2189&0.3763&0.3917&0.5164&0.1719&0.3378&0.7613&0.7932&0.3959&0.5256& 1.2817 & 2.2929 \\
         &768&
         \textbf{0.1705}&\textbf{0.3368}
         &\underline{0.1733}&\underline{0.3404}&0.2999&0.4599&0.3546&0.4922&0.2127&0.3711&0.7940&0.8163&0.3774&0.5125& 1.1972 & 1.8658 \\
         \hline
        
         \multirow{5}{*}{\rotatebox{90}{Wind}}
         &48&
         \textbf{2.6124}&\textbf{1.1886}
         &3.5491&1.4283&\underline{2.7963}&\underline{1.1904}&3.3011&1.2922&3.3916&1.3999&3.0307&1.2898&2.9602&1.2638& 4.0928& 1.8116  \\
         &96&
         \textbf{3.1175}&\textbf{1.3198}
         &4.0628&1.5638&3.3353&\underline{1.3279}&3.5927&1.3374&4.0250&1.5616&\underline{3.2913}&1.3322&3.3277&1.3292 & 3.9678& 1.7949 \\
         &192&
         \textbf{3.3957}&\textbf{1.3623}
         &4.2476&1.5732&3.6808&\underline{1.3659}&3.7467&1.3671&4.2043&1.5623&\underline{3.5763}&1.3773&3.7408&1.3951 & 4.1021 & 1.7965\\
         &384&
         \textbf{3.5119}&\textbf{1.3748}
         &4.3452&1.5886&3.7133&\underline{1.3755}&3.7970&1.3800&4.3374&1.5979&\underline{3.6803}&1.3798&3.7605&1.3815 & 3.9905 & 1.7902\\
         &768&
         \textbf{3.5959}&\textbf{1.3853}
         &4.0653&1.5428&3.7967&1.3860&3.8205&\underline{1.3857}&4.0893&1.5324&\underline{3.7261}&1.3888&3.8065&1.3851& 3.9071 & 1.7978 \\
         \hline
         %
         \multirow{4}{*}{\rotatebox{90}{AirDelay}}
         & 96 & \textbf{0.4687}& \textbf{0.3120}& 0.4809 & 0.3348 & 0.5157 & 0.3954 & 0.4885 & 0.3594 &\underline{0.4722} & \underline{0.3190}& 0.5012 & 0.3870 & 0.4799 & 0.3336& 0.6866 & 0.9315\\
         & 192 & \textbf{0.4727} & \textbf{0.3167} & 0.4887 & 0.3385 & 0.5355 & 0.4214 &0.4848 & 0.3385 & \underline{0.4768} & \underline{0.3266} & 0.5067 & 0.3927 & 0.4874 & 0.3478 & 1.1279 & 1.9009 \\
         & 384 & \textbf{0.4800} & \textbf{0.3176} & 0.5028 & 0.3402 & 0.5563 & 0.4429 & 0.4950 & 0.3530& \underline{0.4820} & \underline{0.3212} & 0.5142 & 0.3966 & 0.4993 & 0.3662  & 0.8286 &1.2476\\
         & 768 & \textbf{0.4894} &\textbf{0.3216} & 0.5193 & 0.3516 & 0.6081 & 0.4895 & 0.5041 & 0.3625& \underline{0.4953} & \underline{0.3418} & 0.5185 & 0.3932 & 0.5071 & 0.3704 &  1.2700&2.2073 \\
         \hline
         \hline
    \end{tabular}
    \label{tab:uni}
    \vspace{-3ex}
\end{table*}

\subsection{{Prediction Results of Multivariate LTTF}}
\vspace{-1ex}

We compare \ourmodel to other baselines in terms of MSE and MAE under the  multivariate time-series forecasting setting, and the results are reported in \cref{tab:multi}. 
We can observe that \ourmodel outperforms SOTA Transformer-based models, as well as other competitive methods, under different predict-length settings. 
For example, 
%
under the predict-96 setting, {\revise{compared to the second best results}}, \ourmodel achieves 
41.0\% (0.2021$\rightarrow$0.1193), 
20.2\% (0.8586$\rightarrow$ 0.6854), 
5.2\% (1.2371$\rightarrow$1.1725) and 
4.1\% (0.7276$\rightarrow$0.6978) 
MSE reductions on Exchange, ETTm1, Wind and ETTh1 datasets,
respectively. 
%
%
Besides, when $L_y=384$, 
\ourmodel achieves 
41.6\% (0.8105$\rightarrow$0.4730), 
33.5\% (0.3976$\rightarrow$0.2643), 
16.3\% (1.1112$\rightarrow$0.9298) and 
9.4\% (0.5518$\rightarrow$0.4997) 
MSE reductions on Exchange, ECL, ETTm1 and Weather datasets, respectively, 
as well as 
28.5\% (0.7513$\rightarrow$0.5369), 
13.5\% (0.4183 $\rightarrow$0.3620) and 
8.0\% (0.7593$\rightarrow$0.6988) 
%
MAE reductions on Exchange, ECL and ETTm1 datasets, respectively. 
%
Moreover,
when the predict-length $L_y$ is prolonged to 768, \ourmodel achieves 
39.9\% (0.5651$\rightarrow$0.3396), 
19.7\% (1.2247$\rightarrow$0.9835) and 
6.0\% (1.4580$\rightarrow$1.3698) 
MSE reductions on ECL, ETTm1 and Wind datasets, respectively, plus 
21.0\% (0.5182$\rightarrow$0.4092) and 
9.4\% (0.7940$\rightarrow$0.7193) 
MAE reductions on ECL and ETTm1 datasets, respectively.

On the other hand, in general, 
the Transformer-based models outperform the RNN-based models. 
This shows the strength of the self-attention mechanism in extracting intricate temporal dependencies in high-dimensional time-series data. 
Moreover, MSE and MAE scores of \ourmodel grows slower as the predict length prolongs than other baselines indicating better stability of our proposed model. 
For the datasets w/ periodicity (\eg Weather, ECL) and w/o periodicity (\eg Exchange), 
\ourmodel consistently delivers good performance, 
which suggests the promising generalization ability. 
{\revise{In addition, for the dataset with irregular time intervals (\eg AirDelay), \ourmodel still achieves the best performance consistently, while the improvements are less significant. 
This suggests that the temporal patterns in less-structured time-series data are more challenging for deep models to capture.}}

{\bf{\revise{Forecasting with Time-Determined Lengths.}}}
\revise{
%
We further evaluate the performance of multivariate LTTF when the input and output lengths are configured as time-determined intervals, \eg 1 day. 
In particular, in this experiment, the input length $L_x$ is set to 1 day and the output length $L_y$ is chosen from $\{ 1 \, \text{day} \, (1\text{D}), $ $1 \, \text{week} \, (1\text{W}), $ $ 2 \, \text{weeks} \, (2\text{W}), $ $ 1 \, \text{month} \, (1\text{M})\}$.  
We inspect forecasting performances of different methods on ETTh1 and ETTm1 datasets. 
The results are reported in~\cref{tab:time_inv}. 
%
As depicted, \ourmodel still achieves the best (or competitive) performance, which suggests the high capacity of \ourmodel in perceiving long-term signals. 
}

\begin{table*}[t]
    \centering
    \small 
    \tabcolsep 0.05in
    \caption{
    Ablation study of the input representation. 
    }
    \label{tab:abl_scale}
    \vspace{-1.5ex}
    \begin{tabular}{c|c|c|c|c|c|c|c|c|c|c}
     \hline
     \hline
     \multicolumn{2}{c|}{Dataset} & \multicolumn{5}{c|}{ECL}&\multicolumn{4}{c}{ETTm1}\\
     \hline
     \multicolumn{2}{c|}{Predict Length } & 48&96&192&384&768&96&192&384&768\\
      \hline
       \multirow{2}{*}{
       $\mathcal{X}^{in} = \mathcal{X}^{v} + \bar{\Gamma}^{S}$ (refer to \cref{eq:H_in})
       } 
       &MSE& 0.1921 & 0.2124 & 0.2378 & 0.2643 & 0.3396 & 0.6954 & 0.7856 & 0.9298 & 0.9835 \\
       &MAE& 0.3034 & 0.3193 & 0.3456 & 0.3620 & 0.4092 & 0.5901 & 0.6387 & 0.6988 & 0.7193 \\
      \hline
      \multirow{2}{*}{
      $\mathcal{X}^{in}_{- \Gamma} \defeq \mathcal{X}^{v}$
      }
       &MSE& 0.1995 & 0.2614 & 0.2787 & 0.2932 & 0.3406 & 0.8342 & 0.9878 & 1.0936 & 1.1578\\
       &MAE& 0.3123 & 0.3659 & 0.3766 & 0.3791 & 0.4083 & 0.6623 & 0.7353 & 0.7786 & 0.8066\\
      \hline
      \multirow{2}{*}{
      $ \mathcal{X}^{in}_{-\mathcal{R}} \defeq 
      \mathbf{W}^{v} \odot \mathcal{X} + \mathbf{b}^{v} + \bar{\Gamma}^{S} $
      }
       &MSE& 0.1896 & 0.2178 & 0.2421 & 0.2674 & 0.3425 & 0.7216 & 0.8313 & 0.9429 & 0.9794\\
       &MAE& 0.3002 & 0.3257 & 0.3512 & 0.3654 & 0.4109 & 0.6107 & 0.6593 & 0.7031 & 0.7152\\
       \hline
       %
       \multirow{2}{*}{
       $\mathcal{X}^{in}_{-\mathcal{R}-\Gamma} \defeq \mathbf{W}^{v} \odot \mathcal{X} + \mathbf{b}^{v}$
       }
       &MSE& 0.2010 & 0.2735 & 0.2749 & 0.3078 & 0.3391 & 0.8853 & 0.9754 & 1.1112 & 1.1538\\
       &MAE& 0.3125 & 0.3770 & 0.3710 & 0.3899 & 0.4076 & 0.6846 & 0.7387 & 0.7842 & 0.7996\\
        \hline
       \multirow{2}{*}{
       $\mathcal{X}^{in}_{-\mathcal{X}} \defeq \mathbf{W}^{v} \odot \mathbf{W}^{\mathcal{R}} \mathcal{X} + \mathbf{b}^{v} + \bar{\Gamma}^{S} $
       }
       & MSE & 0.2774 & 0.2622 & 0.2789  & 0.3040 & 0.3065 &0.8342 & 0.9878 & 1.0936 & 1.1578\\
       &MAE& 0.3638 & 0.3557 & 0.3732 & 0.3955 & 0.3889 & 0.6623 & 0.7353 & 0.7786 & 0.8066\\
       \hline
       \multirow{2}{*}{
       $\mathcal{X}^{in}_{-\mathcal{X}-\Gamma} \defeq \mathbf{W}^{v} \odot \mathbf{W}^{\mathcal{R}} \mathcal{X} + \mathbf{b}^{v}$
       }
       &MSE& 0.2493 & 0.2631 & 0.2649 & 0.2931 & 0.3217 & 0.7344 & 0.8455 & 1.0541 & 1.0540\\
       &MAE& 0.3404 & 0.3473 & 0.3561 & 0.3822 & 0.4057 & 0.6221 & 0.6636 & 0.7540 & 0.7474\\
       \hline
       \hline
    \end{tabular}
    \vspace{-1.25ex}
\end{table*}

\begin{table*}[t]
    \centering
    \small 
    \tabcolsep 0.04in
    \caption{
    Ablation study of the Stationary and Instant Recurrent Network (on Wind dataset).
    }
    \vspace{-1.5ex}
    \begin{tabular}{c|cc|cc|cc|cc|cc|cc}
     \hline
     \hline
     Setting  & \multicolumn{6}{c|}{Multivariate Time-Series Forecasting} & \multicolumn{6}{c}{Univariate Time-Series Forecasting}\\
     \hline
     \multirow{2}{*}{Predict Length}
     &
     \multicolumn{2}{c|}{48}&\multicolumn{2}{c|}{96}&\multicolumn{2}{c|}{192}&\multicolumn{2}{c|}{48}&\multicolumn{2}{c|}{96}&\multicolumn{2}{c}{192}\\
     \cline{2-13}
     &MSE&MAE&MSE&MAE&MSE&MAE&MSE&MAE&MSE&MAE&MSE&MAE\\
      \hline
       \ourmodel (with full SIRN) & \textbf{0.9479}&\textbf{0.6539}&\textbf{1.1725}&\textbf{0.7641}&\textbf{1.3291}&\textbf{0.8464}&\textbf{2.6124}&\textbf{1.1886}&\textbf{3.1175}&\textbf{1.3198}&\textbf{3.3957}&\textbf{1.3623}\\
       \hline
       \ourmodel (with Auto-Corr~\cite{autoformer})
       & 1.0253& 0.7109&1.2878&0.8191&1.4263&0.8742&2.7366&1.2251&3.2173&1.3381&3.4182&1.3816\\
       \ourmodel (with Prob-Attn \cite{Informer})
       & 1.0182&0.7069&1.2817&0.8144&1.4246&0.8734&2.7557&1.2229&3.2231&1.3425&3.4423&1.3801\\
       \ourmodel (with LSH-Attn \cite{reformer}) 
       & 1.0223&0.7086&1.2778&0.8136&1.4209&0.8730&2.7454&1.2249&3.1930&1.3405&3.4140&1.3793\\
       \ourmodel (with Log-Attn \cite{logsparse}) 
       & 1.0393&0.7157&1.2866&0.8165&1.4272&0.8755&2.7449&1.2365&3.2116&1.3476&3.4148&1.3831\\
       \ourmodel (with Full-Attn \cite{attention})
       & 1.0165&0.7070&1.2756&0.8117&1.4195&0.8715&2.7356&1.2229&3.1964&1.3477&3.4165&1.3809\\
       \hline
       \hline
    \end{tabular}
    \label{tab:abl_attention}
    \vspace{-1.25ex}
\end{table*}

\begin{table*}[t]
    \centering
    \tabcolsep 0.04in
    \caption{
    {\revise{Ablation Study of Normalizing Flow for LTTF on the Wind dataset.}}
    }
    \label{tab:abl_nl}
    \vspace{-1.5ex}
    \begin{tabular}{c|cc|cc|cc|cc|cc|cc}
     \hline
     \hline
     Setting
     & 
     \multicolumn{6}{c|}{Multivariate Time-series Forecasting}
     &
     \multicolumn{6}{c}{Univariate Time-series Forecasting}
     \\
     \hline
     \multirow{2}{*}{Predict Length} &
     \multicolumn{2}{|c|}{48}&\multicolumn{2}{c|}{96}&\multicolumn{2}{c|}{192}&\multicolumn{2}{c|}{48}&\multicolumn{2}{c|}{96}&\multicolumn{2}{c}{192} \\
     \cline{2-13}
     &MSE&MAE&MSE&MAE&MSE&MAE&MSE&MAE&MSE&MAE&MSE&MAE\\
     \hline
     \ourmodel 
     & 
     \textbf{0.9479}&\textbf{0.6539}&\textbf{1.1725}&\textbf{0.7641}&\textbf{1.3291}&\textbf{0.8464}&\textbf{2.6124}&\textbf{1.1886}&\textbf{3.1175}&\textbf{1.3198}&\textbf{3.3957}&\textbf{1.3623} \\
     \hline
     \ourmodel$_{-NF}^{\mathbf{z}_e + \mathbf{z}_d}$ 
     &
     1.0082&0.7015&1.2488&0.8017&1.4095&0.8686&2.7961&1.2492&3.3128&1.3807&3.4604&1.4155\\
     \hline
     \ourmodel$_{-NF}^{\mathbf{z}_e}$ 
     &   
     0.9866&0.6953&1.2163&0.7960&1.3632&0.8599&2.7514&1.2363&3.2797&1.3814&3.4432&1.4176\\
     \hline
     \ourmodel$_{-NF}^{\mathbf{z}_d}$ 
     & 
     0.9956&0.6949&1.2167&0.7954&1.3682&0.8473&2.7977&1.2651&3.3767&1.4256&3.5208&1.4302\\
     \hline
     \ourmodel$_{-NF}$ 
     & 
     0.9796 & 0.6927 & 1.2184 & 0.8015 & 1.3455 & 0.8517 & 2.7974 & 1.2614 & 3.5117 & 1.4469 & 3.4421 & 1.4159\\
       \hline
       \hline
    \end{tabular}
    \vspace{-3.5ex}
\end{table*}

\subsection{{Performance Comparisons Under Univariate LTTF}} 
\vspace{-1ex}

\cref{tab:uni} reports {\revise{prediction performances}} of different methods under the univariate LTTF setting. 
{\revise{\ourmodel achieves the best (or competitive) MSE and MAE scores under various predict-length settings.~
}}%
 %
%
In particular, satisfactory prediction improvements {\revise{can be observed}} on Exchange, ECL and Weather datasets.  
{\revise{For instance, compared to the second best results, \ourmodel achieves}} 45.8\% (0.3889$\rightarrow$0.2107) MSE reduction under predict-192 on Exchange dataset, 
and 15.9\% (0.4352$\rightarrow$0.3659) on ECL dataset and   
7.4\%  (0.0914$\rightarrow$0.0846) on Weather dataset     
under predict-384 and predict-48, respectively. 
%
%
\revise{
Moreover, \ourmodel still achieves the best scores on the AirDelay dataset, which further demonstrates the effectiveness of \ourmodel in extracting complex temporal patterns.~%
}%

In addition, under the univariate {\revise{LTTF}} setting, 
we find that RNN-based methods achieve competitive prediction results on Weather and Wind datasets, 
which can validate the advantages of RNN in extracting temporal dynamics of the time-series data with low entropy and regular patterns.

\begin{figure*}[t]
    \centering
    \begin{subfigure}[t]{0.23\textwidth}
	\centering
    \includegraphics[width=0.9\textwidth, center]{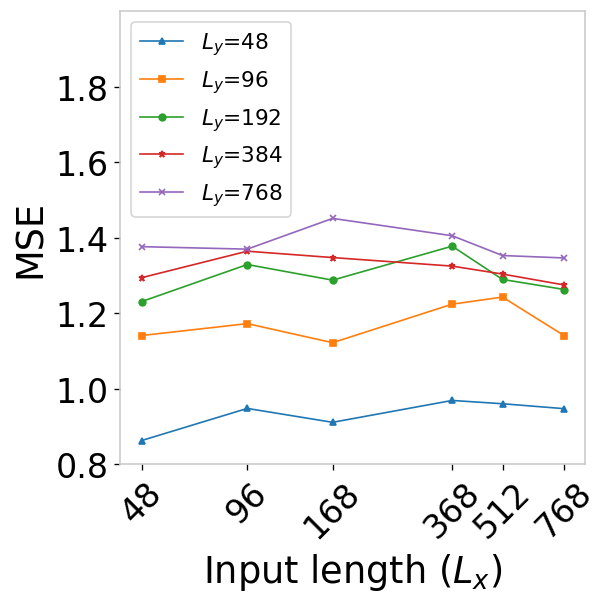}
    \caption{Input length.}
    \label{fig:exp-sensitive-input-length}
	\end{subfigure}
    \begin{subfigure}[t]{0.23\textwidth}
	\centering
    \includegraphics[width=0.9\textwidth, center]{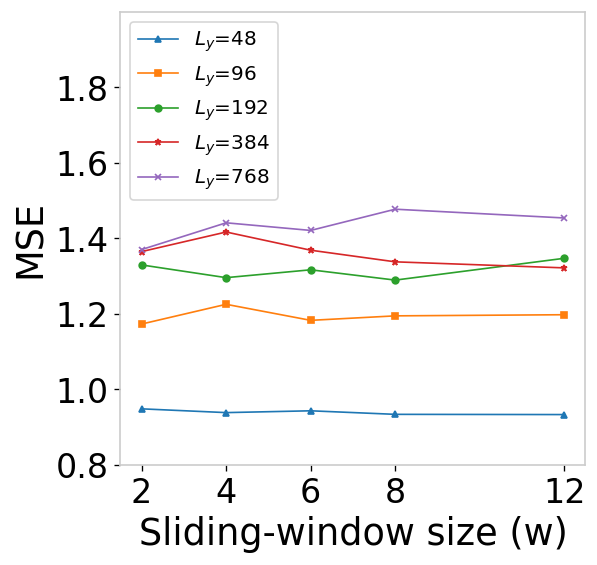}
    \caption{Window size $w$.}
    \label{fig:exp-sensitive-window-size}
	\end{subfigure}
    \begin{subfigure}[t]{0.23\textwidth}
	\centering
    \includegraphics[width=0.9\textwidth, center]{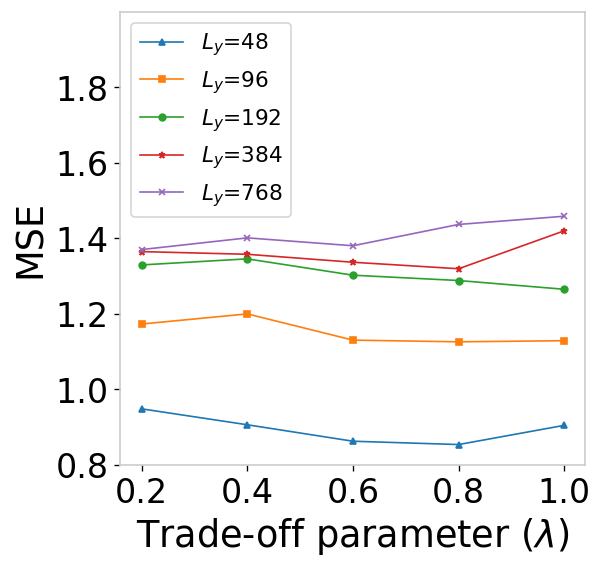}
    \caption{Trade-off parameter $\lambda$.}
    \label{fig:exp-sensitive-loss-weight}
	\end{subfigure}
    \begin{subfigure}[t]{0.23\textwidth}
	\centering
    \includegraphics[width=0.9\textwidth, center]{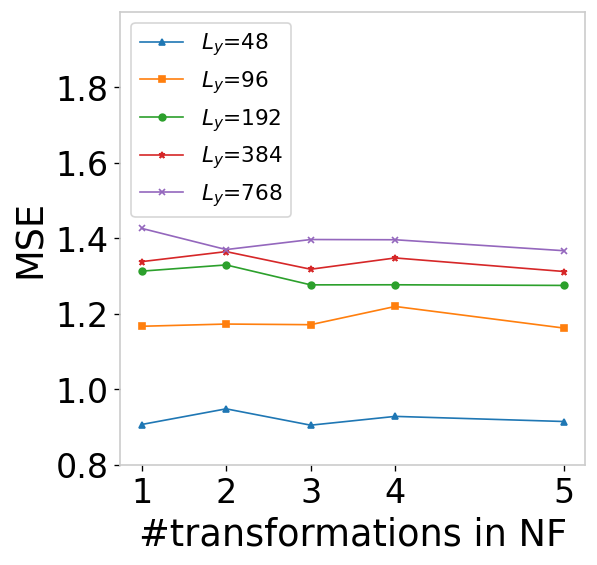}
    \caption{Number of transformations in Normalizing Flow.}
    \label{fig:exp-sensitive-NF}  
	\end{subfigure}
	\vspace{-2ex}
    \caption{
    Parameter sensitivity analysis of \ourmodel.
    }
    \label{fig:exp-sensitive}
    \vspace{-3.5ex}
\end{figure*}

\subsection{Ablation Study}

\noindent
{\revise{We conduct the ablation study under multivariate TF setting}}%
{\footnote{Hereinafter, all the experiments are carried out under the multivariate TF setting by default.}}.

\subsubsection{\bf Multivariate Correlation and Multiscale Dynamics} 
\label{sec:exp:ablation:input-represent}

We compare \ourmodel with its tailored variants w.r.t. the multivariate correlation  and multiscale dynamics, 
and report their prediction performances on ECL and ETTm1 datasets. 
From \cref{tab:abl_scale}, we can obtain several insightful clues on how to embed the input series for LTTF.  
{\bf 1)} $\mathcal{X}^{in}$ v. $\mathcal{X}^{in}_{-\mathcal{R}}$: 
Multivariate correlation contributes less when the dimensions of series is higher (\#dims. of ECL data is much larger than ETTm1 data) or the predict-length is prolonged. 
{\bf 2)} $\mathcal{X}^{in}$ v. $\mathcal{X}^{in}_{-\mathcal{X}-\Gamma}$: 
Temporal dependency is more important for the series with lower dimensions, and for high dimensional time-series, the effectiveness of temporal dependency can be replaced by the inter-series correlation when $L_y$ climbs up. 
{\bf 3)} $\mathcal{X}^{in}_{-\mathcal{R}}$ v. $\mathcal{X}^{in}_{-\mathcal{X}}$ and $\mathcal{X}^{in}_{-\mathcal{R}-\Gamma}$ v. $\mathcal{X}^{in}_{-\mathcal{X}-\Gamma}$:
Multiscale dynamics delivers better performance when being guided by the raw series, which holds regardless of \#dims. of time-series. 
Besides, multivariate correlation contributes more than the raw data for low dimensional time-series. 
{\bf 4)} $\mathcal{X}^{in}_{-\mathcal{R}}$ v. $\mathcal{X}^{in}_{-\mathcal{R}-\Gamma}$ and $\mathcal{X}^{in}_{-\mathcal{X}}$ v. $\mathcal{X}^{in}_{-\mathcal{X}-\Gamma}$:
Multiscale dynamics could harm the performance for LTTF when being equipped with the  multivariate correlation if the raw time-series is absent.

\vspace{-.5ex}
\subsubsection{\bf Stationary and Instant Recurrent Network (SIRN)}

For the ablation study of the proposed SIRN, 
we compare \ourmodel to its different variants by tailoring the encoder-decoder architecture on the Wind dataset, which can be found in \cref{tab:abl_attention}.
Specifically, we replace the sliding-window attention and the RNNs with other self-attention mechanisms to verify the effectiveness of SIRN. 
From \cref{tab:abl_attention}, 
we can see that SIRN achieves best performance under different settings, which validate the effectiveness of information utilization of combining the local and global patterns.

\begin{figure}[t]
    \centering
    \begin{subfigure}[t]{0.235\textwidth}
	\centering
    \includegraphics[width=0.95\textwidth,center]{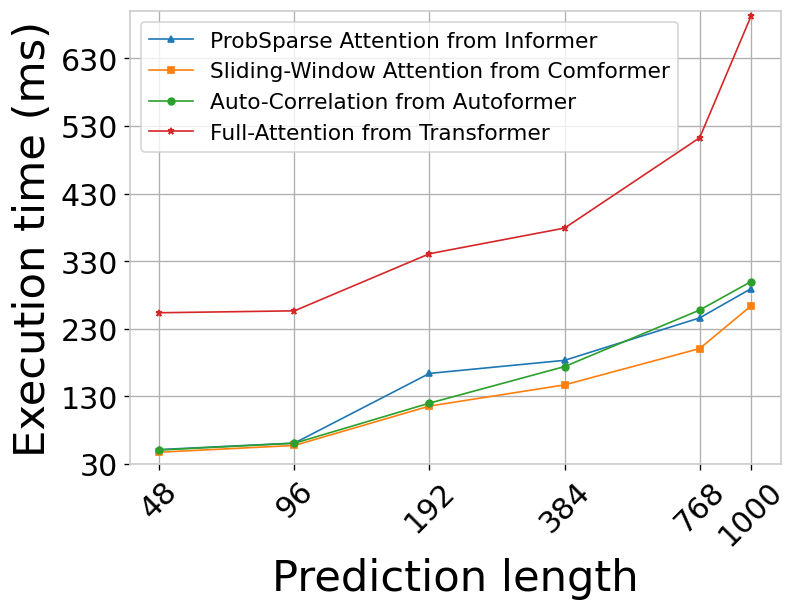}
    \caption{Time complexity.}
    \label{fig:complexity-analysis:time}
	\end{subfigure}
	\begin{subfigure}[t]{0.235\textwidth}
	\centering
    \includegraphics[width=0.95\textwidth,center]{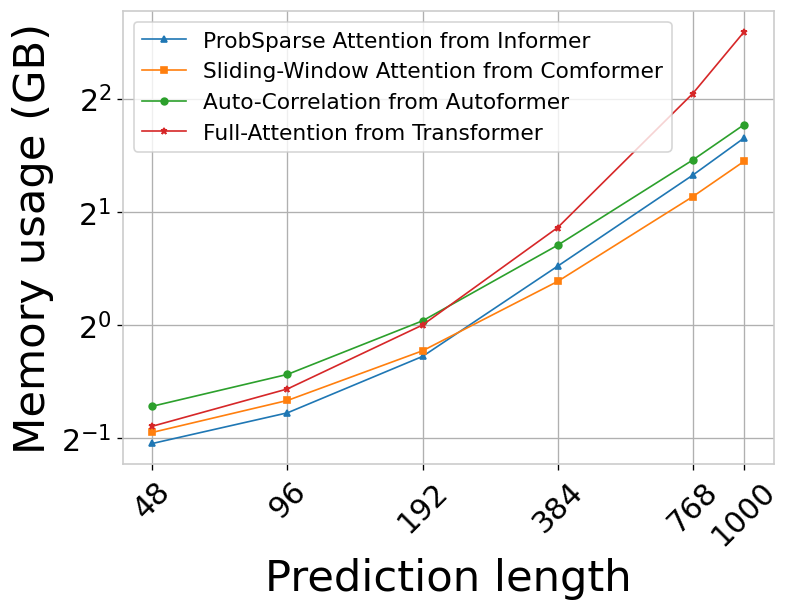}
    \caption{Memory cost.}
    \label{fig:complexity-analysis:memory}
	\end{subfigure}
	\vspace{-1ex}
	\caption{
	Computation efficiency analysis. 
	The input length is set as 96 and all the experiments are conducted on the Wind dataset under the multivariate forecasting setting.
	}
	\label{fig:efficiency-analysis}
	\vspace{-3ex}
\end{figure}

\subsubsection{\bf Normalizing Flow}

%

To verify the effectiveness of normalizing flow block in \ourmodel for LTTF task, 
%
we compare the original \ourmodel with its several variants. 
In particular, we tailor the normalizing flow block in \ourmodel by realizing a generative forecast method with the help of Gaussian probabilistic model as follows: 

\begin{enumerate}[leftmargin=0.5cm,label={$\bullet$}]
    \item 
    \ourmodel$_{-NF}^{\mathbf{z}_e}$: 
    The outcome distribution $\mathbf{z}_{t} $ (yielded by normalizing flow) is replaced by $\mathbf{z}_e$ (obtained by \cref{z_e}). 
    
    \item
    \ourmodel$_{-NF}^{\mathbf{z}_d}$: 
    The outcome distribution $\mathbf{z}_{t} $ (yielded by normalizing flow) is replaced by $\mathbf{z}_d$.
    In particular, we replace $\mathbf{h}_e$ with $\mathbf{h}_d$ in \cref{z_e} and generate $\mathbf{z}_d$ accordingly. 
    
    \item
    \ourmodel$_{-NF}^{\mathbf{z}_e + \mathbf{z}_d}$: 
    The outcome distribution $\mathbf{z}_{t} $ (yielded by normalizing flow) is replaced by $\mathbf{z}_0$ (obtained by \cref{z_0}). 
    
    \item
    \ourmodel$_{-NF}$: 
    We implement a tailored \ourmodel by removing the normalizing flow framework. 
    
\end{enumerate}

The prediction results on Wind dataset are reported in~\cref{tab:abl_nl}. 
We can observe that: 1) The contribution of {\em normalizing flow} is {\revise{indispensable}} for LTTF regardless of forecast setting and predict length, and 2) the way that we adapt the {\em normalizing flow} to the LTTF task is effective.

\begin{table*}[t]
    \centering
    \small 
    \tabcolsep 0.07in
    \caption{
    Comparisons of fusing inter-series correlation and time dependency for LTTF. 
    }
    \vspace{-1.5ex}
    \begin{tabular}{c|c
    |c|c|c|c|c|c|c|c|c}
     \hline
     \hline
     \multicolumn{2}{c|}{
     Dataset 
     } & 
     \multicolumn{5}{c|}{ECL}&\multicolumn{4}{c}{Exchange}\\
      \hline
      \multicolumn{2}{c|}{
      Predict Length 
      } & 48 & 96 & 192 & 384 & 768 & 48 & 96 & 192 & 384 \\
      \hline
      \multirow{2}{*}{\ourmodel} 
       & MSE & {\bf 0.1921} & {\bf 0.2124} & {\bf 0.2378} & {\bf 0.2643} & 0.3396 
       & {\bf 0.0764} & {\bf 0.1193} & 0.2900 & {\bf 0.4730} \\
       &MAE& 0.3034 & {\bf 0.3193} & {\bf 0.3456} & {\bf 0.3620} & 0.4092 & 
       {\bf 0.2093} & {\bf 0.2607} & 0.4187 & {\bf 0.5369} \\
      \hline
      \multirow{2}{*}{\ourmodel (Method 1)}
       &MSE& 0.2003 & 0.2713 & 0.2826 & 0.2898 & 0.3441 & 0.1839 & 0.2938 & 0.4347 & 1.0596 \\
       &MAE& 0.3117 & 0.3784 & 0.3790 & 0.3775 & 0.4150 & 0.3371 & 0.4313 & 0.5190 & 0.8072 \\
      \hline
      \multirow{2}{*}{\ourmodel (Method 2)}
       & MSE & 0.1965 & 0.2354 & 0.2632 & 0.2987 & 0.3437 & 0.1593 & 0.3433 & 0.4321 & 0.5486\\
       & MAE & {\bf 0.3007} & 0.3323 & 0.3587 & 0.3821 & 0.4191 & 0.3144 & 0.4530 & 0.5197 & 0.6007\\
      \hline
      \multirow{2}{*}{\ourmodel (Method 3)}
       &MSE& 0.1997 & 0.2791 & 0.2771 & 0.3061 & 0.3433 & 0.2443 & 0.3310 & 0.4089 & 0.9034\\
       &MAE& 0.3117 & 0.3854 & 0.3744 & 0.3881 & 0.4135 & 0.3795 & 0.4597 & 0.4965 & 0.7444\\
      \hline
      \multirow{2}{*}{\ourmodel (Method 4)}
       &MSE& 0.2010 & 0.2735 & 0.2749 & 0.3078 & {\bf 0.3391} & 0.1135 & 0.1534 & {\bf 0.2344} & 0.5701\\
       &MAE& 0.3135 & 0.3770 & 0.3710 & 0.3899 & {\bf 0.4076} & 0.2597 & 0.3055 & {\bf 0.3805} & 0.5978\\
      \hline
      \hline
    \end{tabular}
    \label{tab:anay_input}
    \vspace{-1.5ex}
\end{table*}

\begin{table*}[t]
    \centering
    \small 
    \tabcolsep 0.07in
    \caption{
    {\revise{Comparisons of feeding hidden states to the {\em normalizing flow} block. The best scores are in boldface and the 2nd best scores are in underlines.}} 
    }
    \label{tab:hstate:nf}
    \vspace{-1.5ex}
    \begin{tabular}{c|c
    |c|c|c|c|c|c|c|c|c}
     \hline
     \hline
     \multicolumn{2}{c|}{
     Dataset 
     } 
     & 
     \multicolumn{5}{c|}{ECL}&\multicolumn{4}{c}{Exchange}\\
      \hline
      \multicolumn{2}{c|}{Predict Length } & 48&96&192&384&768&48&96&192&384\\
      \hline
     \multirow{2}{*}{\ourmodel } 
     & 
     MSE& 0.1921 & {\bf 0.2124} & {\bf 0.2378} & {\bf 0.2643} & \underline{0.3396}
      & {\bf 0.0764} & {\bf 0.1193} & 0.2900 & {\bf 0.4730} \\
      &MAE& 0.3034 & {\bf 0.3193} & {\bf 0.3456} & {\bf 0.3620} & {\bf 0.4092} & 
      {\bf 0.2093} & {\bf 0.2607} & 0.4187 & {\bf 0.5369} \\
      \hline
      \multirow{2}{*}{
      \ourmodel($\mathbf{h}^{(e)}_{k}$, $\mathbf{h}^{(d)}_{k}$) 
      }
       &MSE& {\bf 0.1901} & 0.2300 & 0.2814 & 0.3057 & {\bf 0.3387} & \underline{0.1150} & 0.1506 & 0.2787 & 0.5593\\
       &MAE& {\bf 0.3010} & 0.3322 & 0.3776 & 0.3920 & 0.4168 & \underline{0.2643} & 0.3013 & 0.4108 & 0.5872\\
      \hline
      \multirow{2}{*}{
      \ourmodel($\mathbf{h}^{(e)}_{1}$, $\mathbf{h}^{(d)}_{k}$) 
      }
       &MSE& 0.2004 & 0.2283 & 0.2554 & 0.2896 & 0.3398 & 0.1156 & \underline{0.1476} & \underline{0.2577} & 0.6053\\
       &MAE& 0.3112 & 0.3321 & 0.3588 & 0.3782 & \underline{0.4121} & 0.2655 & \underline{0.2983} & {\bf 0.3911} & 0.6086\\
      \hline
      \multirow{2}{*}{
      \ourmodel($\mathbf{h}^{(e)}_{1}$, $\mathbf{h}^{(d)}_{1}$) 
      }
       &MSE& 0.1984 & 0.2265 & \underline{0.2507} & \underline{0.2751} & 0.3565 & 0.1181 & 0.1669 & {\bf 0.2498} & 0.5300\\
       &MAE& 0.3083 & 0.3304 & \underline{0.3543} & \underline{0.3732} & 0.4226 & 0.2676 & 0.3157 & \underline{0.3925} & 0.5609\\
      \hline
       \multirow{2}{*}{
       \ourmodel($\mathbf{h}^{(e)}_{k}$, $\mathbf{h}^{(d)}_{1}$) 
       }
       &MSE& \underline{0.1904} & \underline{0.2202} & 0.2512 & 0.2799 &0.3547 & 0.1155 & 0.1497 & 0.2846 & \underline{0.5203} \\
       &MAE& \underline{0.3018} & \underline{0.3260} & 0.3586 & 0.3796 & 0.4259 & 0.2654 & 0.3004 & 0.4140 & \underline{0.5549} \\
      \hline
      \hline
    \end{tabular}
    \vspace{-3ex}
\end{table*}

\begin{figure*}[t]
    \centering
    \begin{subfigure}[t]{0.24\textwidth}
	\centering
    \includegraphics[width=0.9\textwidth, center]{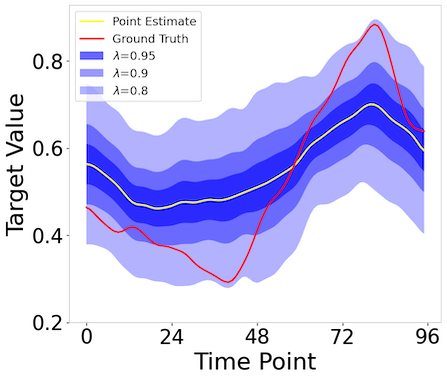}
    \caption{Predict Length = 96.}
    \label{fig:uncertain-ett-96}
	\end{subfigure}
    \begin{subfigure}[t]{0.24\textwidth}
	\centering
    \includegraphics[width=0.9\textwidth, center]{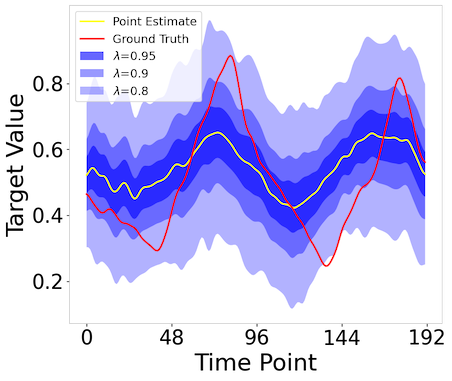}
    \caption{ Predict Length = 192.}
    \label{fig:uncertain-ett-192}
	\end{subfigure}
	\begin{subfigure}[t]{0.24\textwidth}
	\centering
    \includegraphics[width=0.9\textwidth, center]{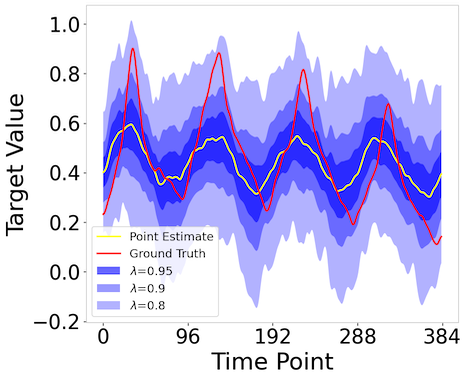}
    \caption{ Predict Length = 384.}
    \label{fig:uncertain-ett-384}
	\end{subfigure}
	\begin{subfigure}[t]{0.24\textwidth}
	\centering
    \includegraphics[width=0.9\textwidth, center]{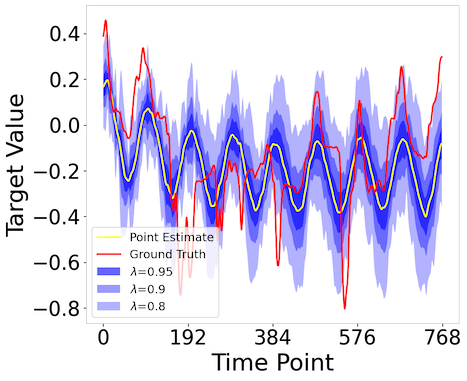}
    \caption{ Predict Length = 768.}
    \label{fig:uncertain-ett-768}
	\end{subfigure}
	\vspace{-1ex}
	\caption{
	With the help of Normalizing Flow, \ourmodel can generate the prediction results with uncertainty quantification for LTTF. 
	Four illustrative cases are demonstrated on {\revise{the}} ETTm1 dataset under the multivariate setting. 
	$\lambda$ denotes the contributions of the encoder-decoder, 
	that is, $1 - \lambda$ represents the  impacts of the normalizing flow block. 
	}
	\label{fig:exp-uncertainty-ett}
	\vspace{-3.5ex}
\end{figure*}

\vspace{-.5ex}
\subsection{Parameter Sensitivity Analysis}
\vspace{-.5ex}

We report {\revise{parameter sensitivity analysis}} in \cref{fig:exp-sensitive}, which is conducted on {\revise{the}} Wind dataset. 
To be specific, we inspect four hyper parameters including the input length $L_{x}$, the window size $w$ of sliding-window attention, the trade-off parameter $\lambda$ and the number of transformations in Normalizing Flow. 
Generally, we can observe that the performance of \ourmodel is quite stable most of the time w.r.t. the varying of different hyper-parameters.
%
In particular, 
as shown in \cref{fig:exp-sensitive-input-length}, 
long-term time-series forecasting setting (\eg $L_y=384$) seems to be more capable of handling longer input, though the volatility of performance is small.   

\vspace{-.5ex}
\subsection{Computational Efficiency Analysis}  \label{sec:exp:complexity}
\vspace{-.5ex}

We conduct execution time consumption and memory usage comparisons between \ourmodel (with sliding-window attention) and other attention mechanisms. 
We replace the standard self-attention mechanism in Transformer with different variants and carry out the prediction with the corresponding method for $10^3$ times (taking the sequences in different time spans as inputs), 
then the averaged running time per forecast is reported. 
For the memory cost comparisons, the maximum memory usage is recorded. 
The time consumption and memory usage of different attentions are demonstrated in \cref{fig:efficiency-analysis}.  
%
\ourmodel performs with better efficiency in both short- and long-term time-series forecasting. 

\begin{figure*}[t]
    \centering
    \begin{subfigure}[t]{0.23\textwidth}
	\centering
    \includegraphics[width=0.9\textwidth, center]{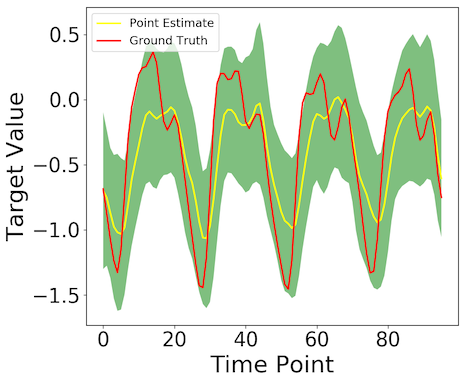}
    \vspace{-3.5ex}
    \caption{ \#transformations = 1.}
	\end{subfigure}
    \begin{subfigure}[t]{0.23\textwidth}
	\centering
    \includegraphics[width=0.9\textwidth, center]{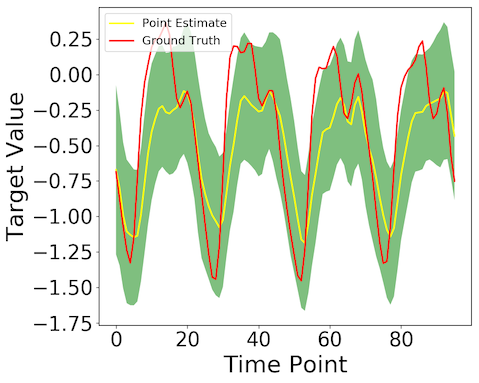}
    \vspace{-3.5ex}
    \caption{ \#transformations = 2.}
	\end{subfigure}
	\begin{subfigure}[t]{0.23\textwidth}
	\centering
    \includegraphics[width=0.9\textwidth, center]{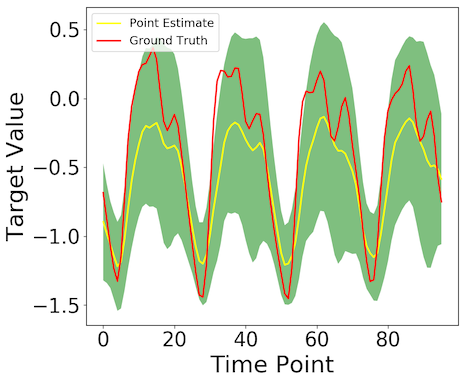}
    \vspace{-3.5ex}
    \caption{ \#transformations = 3.}
	\end{subfigure}
	\begin{subfigure}[t]{0.23\textwidth}
	\centering
    \includegraphics[width=0.9\textwidth, center]{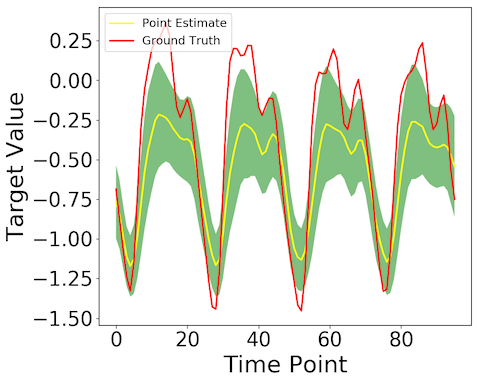}
    \vspace{-3.5ex}
    \caption{ \#transformations = 4.}
    \end{subfigure}
    
    \begin{subfigure}[t]{0.23\textwidth}
	\centering
    \includegraphics[width=0.9\textwidth, center]{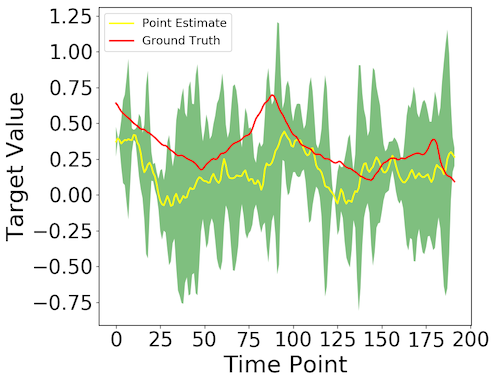}
    \vspace{-3.5ex}
    \caption{ \#transformations = 1.}
	\end{subfigure}
    \begin{subfigure}[t]{0.23\textwidth}
	\centering
    \includegraphics[width=0.9\textwidth, center]{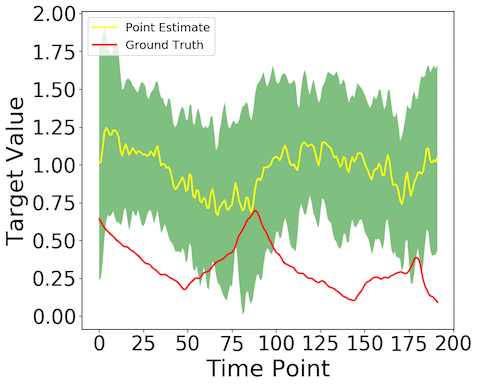}
    \vspace{-3.5ex}
    \caption{ \#transformations = 2.}
	\end{subfigure}
	\begin{subfigure}[t]{0.23\textwidth}
	\centering
    \includegraphics[width=0.9\textwidth, center]{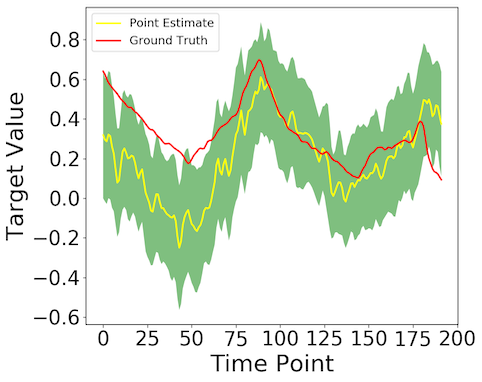}
    \vspace{-3.5ex}
    \caption{ \#transformations = 3.}
	\end{subfigure}
	\begin{subfigure}[t]{0.23\textwidth}
	\centering
    \includegraphics[width=0.9\textwidth, center]{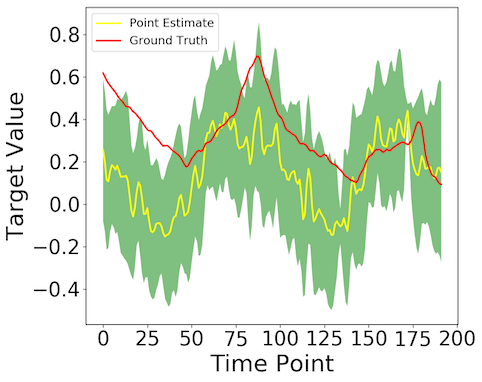}
    \vspace{-3.5ex}
    \caption{ \#transformations = 4.}
	\end{subfigure}
	\vspace{-1ex}
	\caption{
	Uncertainty-aware LTTF with varying \#transforms. 
	To evaluate the performance of {\em normalizing flow} more clearly, we omit the contribution of SIRN by setting $\lambda = 0$ in \cref{eq:loss}. 
	(a)--(d) and (e)--(h) demonstrate two cases in ECL and ETTm1 datasets, respectively. 
	}
	\label{fig:eva-nf}
	\vspace{-1ex}
\end{figure*}

\begin{figure*}[t]
    \centering
    \begin{subfigure}[t]{0.23\textwidth}
	\centering
    \includegraphics[width=0.9\textwidth, center]{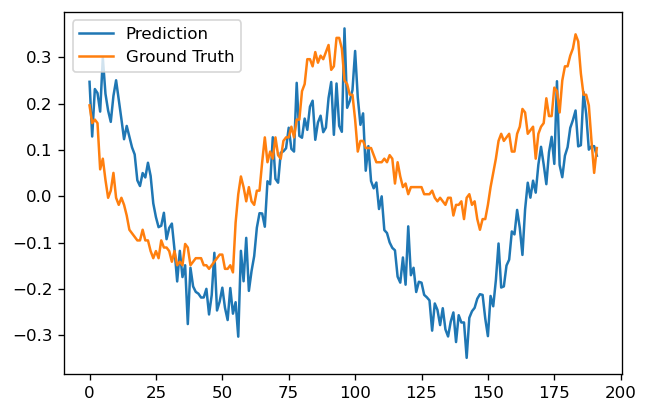}
    \caption{\ourmodel}
    \label{fig:appendix-case-comformer-ett}
	\end{subfigure}
	\begin{subfigure}[t]{0.23\textwidth}
	\centering
    \includegraphics[width=0.9\textwidth, center]{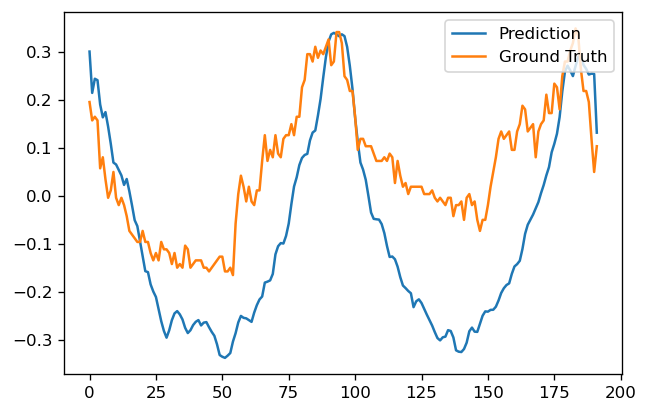}
    \caption{Longformer}
    \label{fig:appendix-case-reformer-ett}
	\end{subfigure}
    \begin{subfigure}[t]{0.23\textwidth}
	\centering
    \includegraphics[width=0.9\textwidth, center]{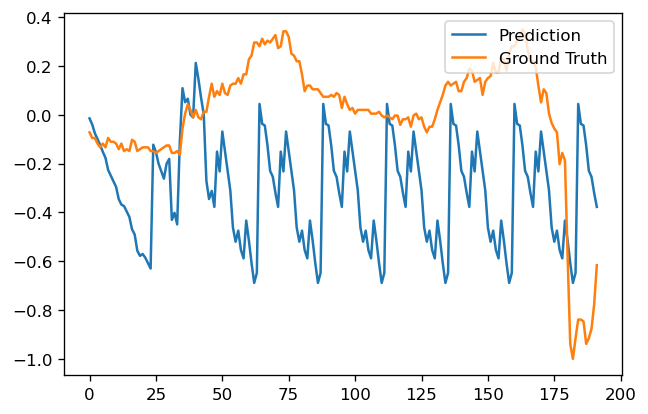}
    \caption{Reformer}
    \label{fig:appendix-case-reformer-ett}
	\end{subfigure}
    \begin{subfigure}[t]{0.23\textwidth}
	\centering
    \includegraphics[width=0.9\textwidth, center]{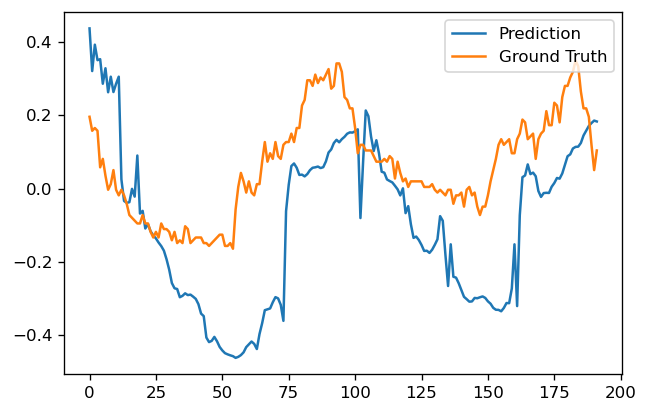}
    \caption{Informer}
    \label{fig:appendix-case-informer-ett}
	\end{subfigure}
    \begin{subfigure}[t]{0.23\textwidth}
	\centering
    \includegraphics[width=0.9\textwidth, center]{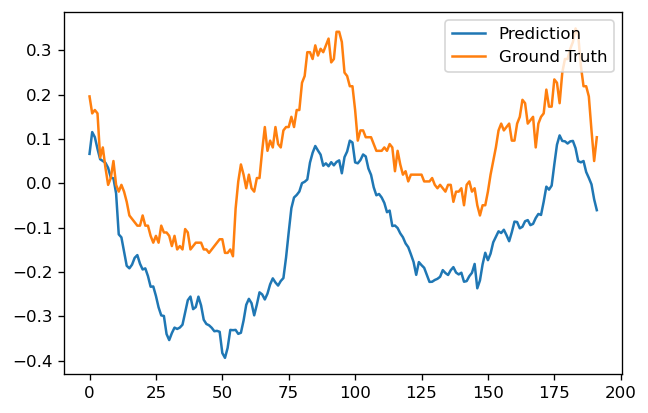}
    \caption{Autoformer}
    \label{fig:appendix-case-autoformer-ett}
	\end{subfigure}
	\begin{subfigure}[t]{0.23\textwidth}
	\centering
    \includegraphics[width=0.9\textwidth, center]{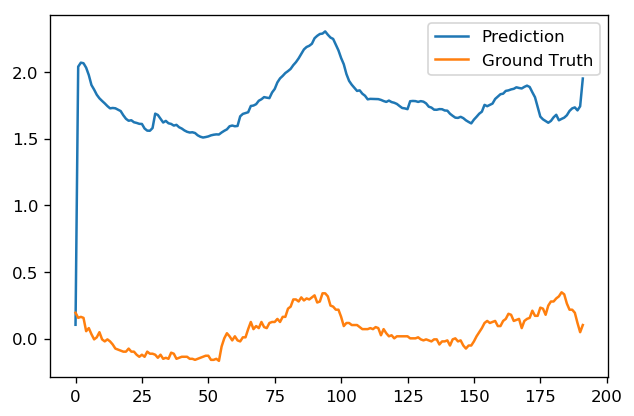}
    \caption{N-Beats}
    \label{fig:appendix-case-nbeats-ett}
	\end{subfigure}
    \begin{subfigure}[t]{0.23\textwidth}
	\centering
    \includegraphics[width=0.9\textwidth, center]{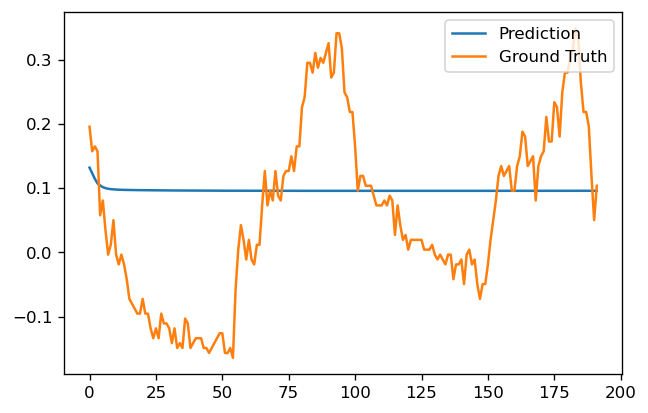}
    \caption{LSTNet}
    \label{fig:appendix-case-lstnet-ett}
	\end{subfigure}
    \begin{subfigure}[t]{0.23\textwidth}
	\centering
    \includegraphics[width=0.9\textwidth, center]{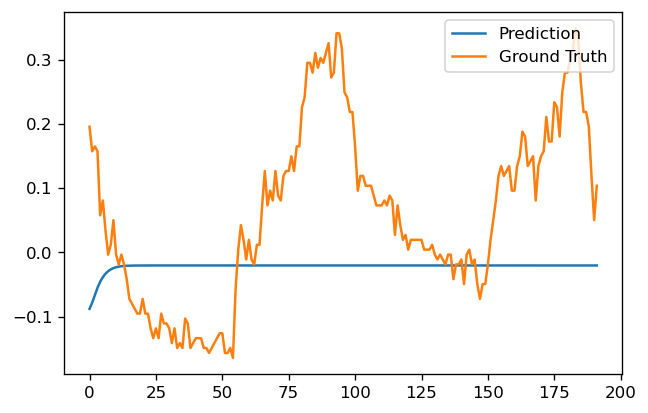}
    \caption{GRU}
    \label{fig:appendix-case-lstm-ett}
	\end{subfigure}
	\vspace{-.5ex}
    \caption{
    Prediction cases on the ETTm1 dataset under the input-96-predict-192 setting. 
    }   \label{fig:showcase-ettm1}
    \vspace{-3.5ex}
\end{figure*}

\vspace{-.5ex}
\subsection{Model Analysis}
\vspace{-.5ex}

\subsubsection{\bf 
Fusing Inter-Series and Across-Time Dependencies%
}

As introduced in \cref{sec:input}, the series data is embedded and fused by taking the multivariate correlation and multiscale dynamics into account. 
To further assess the effectiveness of input representation module in \ourmodel, 
we realize different ways of fusing multivariate correlation and multiscale dynamics {\revise{below}} (let $\mathbf{W}^{\Gamma} = \text{Softmax}(\bar{\Gamma}^{S})$): 

\begin{enumerate}[leftmargin=0.8cm,label={$\bullet$}]
\item 
Method 1: 
$\mathcal{X}^{in} = \mathbf{W}^{v} \odot (\mathbf{W}^{\Gamma} \mathbf{W}^{\mathcal{R}} \mathcal{X} + \mathcal{X}) + \mathbf{b}^{v} $
\item 
Method 2: 
$\mathcal{X}^{in} = \mathbf{W}^{v} \odot (\mathbf{W}^{\mathcal{R}} \mathcal{X} + \mathbf{W}^{\Gamma} \mathcal{X}) + \mathbf{b}^{v} $
\item 
Method 3: 
$\mathcal{X}^{in} = \mathbf{W}^{v} \odot (\mathbf{W}^{\mathcal{R}} \mathcal{X} + \mathbf{W}^{\Gamma} \mathcal{X} + \mathcal{X}) + \mathbf{b}^{v} $
\item 
Method 4: 
$\mathcal{X}^{in} = [\mathbf{W}^{v} \odot (\mathbf{W}^{\mathcal{R}} \mathcal{X} +  \mathcal{X}) + \mathbf{b}^{v}] \mathbf{W}^{\Gamma}$
\end{enumerate}

{\revise{The results}} are reported in~\cref{tab:anay_input}. 
%
We can see that how to fuse the multivariate correlation and temporal dependency is important for {\revise{the}} LTTF task.  
{\revise{This}} impact weighs more for {\revise{low dimensional time-series data}} since the self-attention mechanism in {\revise{encoder-decoder architecture}} can better explore {\revise{intricate dependencies}} when the dimensionality grows.


\subsubsection{\bf 
Uncertainty-Aware Forecasting%
}

The outcome variance of the Normalizing Flow block can suggest the fluctuation range of the forecasting results. 
We randomly select a case in ETTm1 dataset under the multivariate setting and demonstrate the forecasting results with uncertainty quantification for different output lengths in \cref{fig:exp-uncertainty-ett}. 
We can see that \ourmodel tends to make a conservative forecast and the uncertainty quantification can cover the extreme ground truth values if the NF block can be weighted more.

\subsubsection{\bf How Far The Message Should Be Cascaded in Normalizing Flow}

We inspect how the {\em normalizing flow} works for LTTF by varying the number of transformations on two cases in ECL and ETTm1 datasets, respectively, in \cref{fig:eva-nf}.
We can see that the further the latent variable being transformed the better the outcome series performs. 
Therefore, 
the power of {\em normalizing flow} in \ourmodel for LTTF should be explored more dedicatedly.

\subsubsection{\bf How to Feed Hidden States to The Normalizing Flow Block in \ourmodel}

As shown in \cref{fig:frame}, in both encoder and decoder, the first outcome hidden state of the last SIRN layer is fed to the {\em normalizing flow}. 
To assess the effect of feeding hidden states to {\em normalizing flow}, we implement \ourmodel by combining the outcome hidden states in the first/last SIRN layer of the encoder/decoder, which results in 
\ourmodel ($\mathbf{h}^{(e)}_{k}$, $\mathbf{h}^{(d)}_{k}$), 
\ourmodel ($\mathbf{h}^{(e)}_{1}$, $\mathbf{h}^{(d)}_{k}$), 
\ourmodel ($\mathbf{h}^{(e)}_{1}$, $\mathbf{h}^{(d)}_{1}$) and 
\ourmodel ($\mathbf{h}^{(e)}_{k}$, $\mathbf{h}^{(d)}_{1}$) 
where $k$ denotes the last SIRN layer. 
%
We report the prediction results in \cref{tab:hstate:nf}. 
As can be seen, the impact of feeding different hidden states to {\em normalizing flow} is generally marginal though, the low dimensional time-series forecasting is more sensitive to the way of absorbing hidden states for {\em normalizing flow}.

\subsection{Multivariate Time-series Forecasting Showcase}
\vspace{-1ex}

We additionally plot the prediction and the ground truth of the target value.  
The qualitative comparisons between \ourmodel and other baselines on ETTm1 dataset are demonstrated in \cref{fig:showcase-ettm1}. 
We can see that, our model obviously achieves the best performance among different methods.

\subsection{Discussion}
\vspace{-1ex}

{\revise{\bf Windowed Attention: \ourmodel v. Swin Transformer.}}
{\revise{%
The windowed attention mechanism is applied in many applications thanks to its linear complexity, such that the powerful self-attention can be scaled up to large data. 
The very recent Swin Transformer~\cite{liu2021Swin} and its variant~\cite{liu2021swinv2} adopt the windowed attention and devise a shifted window attention to implement a general purpose backbone for computer vision tasks. 
Basically, both \ourmodel and Swin Transformer exploit the self-attention within neighbored/partitioned windows regarding the computational efficiency. 
Besides the locality, connectivity is another merit one can not neglect. 
To achieve connectivity, a shifted window mechanism is proposed for Swin Transformer, while we propose SIRN for \ourmodel so as to absorb long-range dependencies in the time-series data.~%
}}

{\revise{\bf Comparisons of Computational Complexity.}}
{\revise{The windowed attention contributes most to the complexity reduction of \ourmodel. 
Hence, we take different SOTA attention mechanisms as competitors to conduct the computational complexity analysis in \cref{sec:exp:complexity}. 
The computational costs of other components in \ourmodel are not elaborated, which will be provided in our future work. 
}}

\section{Conclusion}

In this paper, we proposed a transformer-based model, namely \ourmodel, to address the long-term time-series forecasting (LTTF) problem. 
Specifically, \ourmodel first embeds the input time series with the multivariate correlation modeling and multiscale dynamics extraction to fuel the downstream self-attention mechanism.
Then, to reduce the computation complexity of self-attention and fully distill the series-level temporal dependencies without sacrificing information utilization for LTTF, 
sliding-window attention, as well as a proposed stationary and instant recurrent network (SIRN),  are equipped to the \ourmodel. 
Moreover, a normalizing flow framework is employed to further absorb the latent states in the SIRN, such that the underlying distribution can be learned  
and the target series can be directly reconstructed in a generative way. 
Extensive empirical studies on six real-world datasets validate  that \ourmodel achieves state-of-the-art performance on long-term time-series forecasting under  multivariate and univariate prediction settings. 
In addition, with the help of normalizing flow, \ourmodel can generate the prediction results with uncertainty quantification.

\vspace{-.5ex}

\section*{Acknowledgment}

\vspace{-.5ex}

We thank China Longyuan Power Group Corp. Ltd. for supporting this work. 
Besides, this work was supported in part by National Key R\&D Progamm of China (No. 2021ZD0110303).

\bibliographystyle{IEEEtran}
\bibliography{reference}

\end{document}